\documentclass[twoside]{article}

\usepackage[accepted]{aistats2022}
\usepackage[round]{natbib}

\renewcommand{\cite}{\citep}

\usepackage{xr-hyper}
\usepackage[colorlinks=true]{hyperref}
\usepackage[dvipsnames]{xcolor}
\hypersetup{
    allcolors = MidnightBlue,
}

\usepackage{bm} 
\usepackage{float} 
\usepackage{graphicx}
\usepackage[linesnumbered,ruled]{algorithm2e}
\usepackage{amsfonts}

\begin{document}

\newcommand{\mc}{\mathcal}

%\runningtitle{Fast and Scalable Spike and Slab Variable Selection in High-Dimensional Gaussian Processes}

% command to supply a shorter version of the authors names so thatg

\twocolumn[

\aistatstitle{Fast and Scalable Spike and Slab Variable Selection in High-Dimensional Gaussian Processes}

\aistatsauthor{ Hugh Dance \And Brooks Paige}

\aistatsaddress{ University College London \And  University College London} ]

\begin{abstract}
       Variable selection in Gaussian processes (GPs) is typically undertaken by thresholding the inverse lengthscales of automatic relevance determination kernels, but in high-dimensional datasets this approach can be unreliable. A more probabilistically principled alternative is to use spike and slab priors and infer a posterior probability of variable inclusion. However, existing implementations in GPs are very costly to run in both high-dimensional and large-$n$ datasets, {\color{black}or are only suitable for unsupervised settings with specific kernels.} As such, we develop a fast and scalable variational inference algorithm for the spike and slab GP that is tractable with arbitrary differentiable kernels. We improve our algorithm's ability to adapt to the sparsity of relevant variables by Bayesian model averaging over hyperparameters, and achieve substantial speed ups using zero temperature posterior restrictions, dropout pruning and nearest neighbour minibatching. In experiments our method consistently outperforms vanilla and sparse variational GPs whilst retaining similar runtimes (even when $n=10^6$) and performs competitively with a spike and slab GP using MCMC but runs up to $1000$ times faster.
\end{abstract}

\section{INTRODUCTION}
Gaussian processes (GPs) are a powerful Bayesian nonparametric approach to regression used widely in machine learning and statistics. GPs enable the user to directly model the distribution over a function $f(\cdot)$ they aim to approximate. The covariance function $k(x,x')$ of the GP enables properties such as stationarity and smoothness to be encoded directly, and flexible function classes can be represented with few (hyper)parameters \cite{rasmussen2003gaussian}. 

To infer variable relevance in GPs `automatic relevance determination' (ARD) kernels are often implemented, which use per-dimension inverse lengthscale parameters $\{\theta_j\}_{j=1}^d$ as a relevance measure. Learning  $\{\theta_j\}_{j=1}^d$ by maximising the marginal likelihood (ML-II) often sees $\theta_j \to 0$ for irrelevant dimensions due to the marginal likelihood's intrinsic complexity penalty. This makes it trivial to select relevant variables by thresholding $\{\theta_j\}_{j=1}^d$ \cite{rasmussen2003gaussian}. However, we find this shrinkage does not hold sufficiently in high-dimensional input designs, {\color{black} making it difficult to perform variable selection reliably}. This agrees with recent evidence \cite{ober2021promises,mohammed2017over} that the marginal likelihood can overfit in heavily parameterised regimes.

An alternative approach is to place spike and slab priors on the inverse lengthscales and infer a per-variable posterior inclusion probability (PIP). Existing implementations in GPs perform excellently but either use Markov Chain Monte Carlo (MCMC) schemes which are very costly to run in high-dimensional or large-$n$ datasets \cite{savitsky2011variable}, {\color{black}or are only suitable for latent input selection in the unsupervised setting (and when using specific kernels) \cite{dai2015spike}}.

\textbf{Our Contribution}: We develop a fast and scalable variational inference algorithm for GPs with spike and slab variable selection priors, which we call the spike and slab variational Gaussian process (SSVGP). By using a continuous approximation to the Dirac spike and a structured variational approximation to the posterior, we are able to derive an otherwise intractable approximate co-ordinate ascent variational inference (CAVI) \cite{blei2017variational} algorithm with similar complexity to ML-II training. We subsequently improve our algorithm's ability to adapt to the sparsity in the data by Bayesian model averaging over spike and slab hyperparameter values. We achieve substantial speed-ups and $\mc O(nf(n)d)$ complexity with $f(n) \in [\log(n),n]$ by using a combination of zero temperature posterior restrictions, dropout pruning, and nearest neighbour minibatching. Our method can be used to improve GP regression predictive accuracy in high-dimensional scenarios, or enable probabilistically principled variable selection through inference on the PIPs - without the usual spike and slab implementation costs.

We find across a range of experiments that our SSVGP consistently outperforms standard GP regression and sparse variational GP regression \cite{titsias2009variational, hensman2013gaussian} in terms of predictive accuracy whilst keeping similar runtimes. Our method also outperformed benchmark variable selection algorithms and matched the performance of a spike and slab GP with MCMC \cite{savitsky2011variable}, but can run up to 1000 times faster. Code for our method is provided at the Github repository {\color{blue}https://github.com/HWDance/SSVGP}.

\section{BACKGROUND AND RELATED WORK}
In this section we review GP regression, spike and slab priors, {\color{black} variational inference and implementation challenges with spike and slab GPs, as well as related work.}

\subsection{Gaussian Process Regression}
A Gaussian process is a random function $f \sim \mc {GP}(m(\bm x),k_{\bm \alpha}(\bm x,\bm x'))$ with mean function $m(\cdot) : \mc X \to \mathbb R$ and positive semi-definite covariance function (`kernel') $k(\cdot, \cdot) : \mc X \times \mc X \to \mathbb R$, such that any finite subset of points are multivariate Gaussian: $\bm f = [f(\bm x_1),...,f(\bm x_n)] \sim \mc N(\bm m, K_{XX})$, $[\bm m]_i = m(\bm x_i)$, $[K_{XX}]_{ij} = k_{\bm \alpha}(\bm x_i,\bm x_j)$. Here $\bm \alpha$ are kernel parameters.

Given a dataset $\mc D = (\bm y,X) \in \mathbb R^n \times \mathbb R^{n \times d}$, we define a Gaussian process model as follows
\begin{align}
\bm y|\bm f & \sim p(\bm y|\bm f)\label{m0}\\
f(\cdot)|\bm \alpha & \sim \mc {GP}(m(\bm x) \label{m1}, k_{\bm \alpha}(\bm x,\bm x'))
\end{align}
In standard GP regression (GPR), the observation model is a Gaussian centred at $\bm f$: $p(\bm y|\bm f) = \prod_{i=1}^n\mc N(y_i|f_i,\sigma^2)$. In this case the marginal likelihood has closed form $p_{\bm \alpha,\sigma}(\bm y) = \mc N(\bm y |\bm m, K_{XX}+\sigma^2I)$ and can be maximised to learn kernel parameters $\bm \alpha$ and noise variance $\sigma^2$ (aka ML-II optimisation). The predictive distribution at any $\bm x_* \in\mathbb R^d$ is also Gaussian with closed form moments \cite{rasmussen2003gaussian}.

 To do variable selection in GPR, automatic relevance determination (ARD) kernels \cite{mackay1998introduction} are traditionally used. {\color{black}Most popular ARD kernels (e.g. squared exponential (SE), Mat\'{e}rn, Cauchy) fall into a subclass of anisotropic stationary kernels of the form
\begin{align}
k_{\tau, \bm \theta}(\bm x, \bm x') = \tau h(||\bm \theta \odot (\bm x - \bm x')||_2).\label{kernel}
\end{align} 
Here $\bm \theta \in \mathbb R_{+}^d$ are per-dimension inverse lengthscales, which are used a measure of variable relevance (assuming inputs are scaled to have unit variance), $\tau \in \mathbb R_{>0}$ is the scale parameter of the kernel and $h(\cdot)$ is a continuous (often monotonically decreasing) function. %One motivation for interpreting $\bm \theta$ as a relevance measure arises through the property that the expected squared function variation in the prior induced by varying $x_j$, is monotonically increasing in $\theta_j\geq0$ if $h'(\cdot)\leq 0$:
%\begin{align*}
%    V_j & = \mathbb E_{p(f)}[(f(\bm x + \bm \epsilon_j)-f(\bm x))^2] \\
%    & = 2\tau - k(\bm x+\bm \epsilon_j,\bm x)
%    \\
%    \frac{\partial V_j}{\partial \theta_j} & = \underbrace{\frac{\partial V_j}{\partial k}}_{<0}\underbrace{\frac{\partial k}{\partial h}}_{\geq 0}\underbrace{\frac{\partial h}{\partial \theta_j}}_{\leq 0} \geq 0
%\end{align*}
}ML-II optimisation often sees $\theta_j \to 0^+$ for irrelavent dimensions due to the marginal likelihood's intrinsic complexity penalty \cite{rasmussen2003gaussian}, and so variable selection can be undertaken by hard-thresholding $\{\theta_j\}_{j=1}^d$. {\color{black}However, we demonstrate that in high-dimensional designs inferring the appropriate threshold can be challenging, and failing to threshold reliably can substantially reduce predictive accuracy. 

We note alternative measures of variable relevance have been developed using posterior predictive sensitivity \cite{paananen2019variable}, however these approaches suffer the same drawback of needing to infer an appropriate threshold for binary selection decisions}.

\subsection{Spike and Slab Priors}
Spike and slab priors \cite{mitchell1988bayesian} are considered a gold standard for high-dimensional variable selection problems due to their excellent properties \cite{narisetty2014bayesian,castillo2015bayesian}, empirical performance and interpretability. Given a model $p(\bm y|\bm \theta)$ with per-dimension `relevance' parameters $\bm \theta \in \Theta  \subset \mathbb R^d$, a spike and slab prior augments the model with binary inclusion variables $\bm \gamma \in \{0,1\}^d$ 
{{\begin{align}p(\bm \theta, \bm \gamma) = \prod\nolimits_{j=1}^d[\mc P_{slab}(\theta_j)\pi]^{\gamma_j}[\delta_{0}(\theta_j)(1-\pi)]^{1-\gamma_j}\end{align}}}Here $\mc P_{slab}(\cdot)$ is the slab distribution (usually set to be approximately uniform over dom($\theta_j$)), the spike distribution is a Dirac point mass at zero, and $\pi$ is the prior probability of inclusion. Since $p(\theta_j = 0|\gamma_j = 0)=1$, variable relevance can be inferred through the posterior inclusion probability (PIP) $p(\gamma_j = 1|\bm y)$ for input dimension $j$
\begin{align*}p(\gamma_j = 1|\bm y) = \frac{\sum_{\bm \gamma_{\neg j}}\int_{\bm \theta}p(\bm y|\bm \theta)p(\bm \theta,\bm \gamma)}{\sum_{\bm \gamma}\int_{\bm \theta}p(\bm y|\bm \theta)p(\bm \theta,\bm \gamma)}\end{align*}
The posterior mode $\bm \gamma^* = \text{argmax}_{\bm \gamma \in \{0,1\}^d}\{p(\bm \gamma|\bm y)\}$ can also be used for variable selection (if available), since this by definition selects the posterior likeliest model. Unfortunately the posterior $p(\bm \gamma|\bm y)$ is generally intractable but can be approximated using Markov Chain Monte Carlo (MCMC) \cite{brooks2011handbook} or variational inference \cite{blei2017variational}, of which the latter we briefly review below.

{\color{black}
\subsection{Variational Inference and Intractabilities in Spike and Slab GPs}

Given a generative model $p(\mc D|\mc Z)p(\mc Z)$ over latent variables $\mc Z$ and data $\mc D$, variational inference can be used to approximate the posterior $p(\mc Z|\mc D)$ (if intractable) with a surrogate $q(\mc Z) \in \mc Q$, where $\mc Q$ is restricted by parametric or independence assumptions. The optimal $q^* \in \mc Q$ is chosen by minimising the Kullback-Leibler divergence $KL[q(\mc Z)||p(\mc Z|\mc D)]$, which is equivalent to maximising the evidence lower bound (or Free Energy) $\langle \log p(\mc D|\mc Z) \rangle_{q(\mc Z)} - KL[q(\mc Z)||p(\mc Z)]$ \cite{blei2017variational}. 

Variational inference appears a natural choice to achieve fast and scalable inference in GPs with spike and slab priors. Factorising the posterior over dimensions $q(\bm \theta, \bm \gamma) = \prod_{j=1}^dq(\theta_j,\gamma_j)$ reduces the variable selection problem from a $2^d$ search problem to optimising $\mc O(d)$ variational parameters (after optimisation marginal PIPs and the posterior mode can be recovered trivially due to the factorisation).  However, even in standard GPR models, optimisation remains either challenging or entirely intractable depending on the independence assumptions used. To see this, consider the evidence lower bound induced for a GPR model with Gaussian marginal likelihood $p(\bm y|\bm \theta) = \mc N(\bm y|\bm m,K_{XX}(\bm \theta)+\sigma^2I)$, spike and slab prior $p(\bm \theta, \bm \gamma)$ and variational approximation $q(\bm \theta, \bm \gamma)$
{{\begin{align*}\mc F & = \langle \log p(\bm y|\bm \theta) \rangle_{q(\bm \theta|\bm \gamma)q(\bm \gamma)}-KL[q(\bm \theta,\bm \gamma)||p(\bm \theta, \bm \gamma)]\end{align*}}}%

If keeping ($\theta_j,\gamma_j$) coupled such that $q(\bm \theta, \bm \gamma) = \prod\nolimits_jq(\theta_j,\gamma_j)$, but $q(\theta_j,\gamma_j) \neq q(\theta_j)q(\gamma_j)$ 
--- i.e.\ the ``paired mean field" (PMF) approximation of \citet{carbonetto2012scalable} ---
then the expected marginal likelihood term is intractable as it requires computing the expected kernel inverse $\langle K^{-1} \rangle_{q(\theta,\gamma)}$. 
Due to the dependence of this term on $q(\bm \gamma)$, only Black Box variational Inference (BBVI) can be used to unbiasedly approximate its gradient \cite{ranganath2014black}, which suffers from high variance. On the other hand, if using a fully factorised (mean field) approximation $q(\bm \theta, \bm \gamma) = \prod\nolimits_jq(\theta_j)q(\gamma_j)$, then the KL term is undefined for any $q(\theta_j) \neq \delta_0(\theta_j)$ due to the need to compute $\langle \log \delta_0(\theta_j) \rangle_{q(\theta_j) \neq \delta_0(\theta_j)}$.
}

\subsection{Related Work on Spike and Slab GPs}

Several previous works have used MCMC for inference with spike and slab priors in GPs \cite{linkletter2006variable,savitsky2011variable,qamar2014additive}, {\color{black} motivated by its asymptotic convergence guarantees}. Whilst performance is typically excellent, these algorithms are costly to run both in large-$d$ datasets as they stochastically search over a $2^d$ model space, and in large-$n$ datasets they use many model evaluations with $\mc O(n^3)$ cost. For example, \citet{savitsky2011variable} report runtimes of nearly $3$ hours when $(n=100, d=1000)$. {\color{black} Although scalable MCMC schemes have since been developed for GPR using Hamiltonian Monte Carlo \cite{hensman2015mcmc,rossi2021sparse}, these schemes require differentiable random variables and so can't be used off-the-shelf with spike and slab priors without bespoke augmentation strategies \cite{pakman2013auxiliary}. The need to store a large number of $d$-dimensional Monte Carlo samples also makes MCMC unattractive for high-dimensional scenarios.
}

%Dai, Hensman and Lawrence (2015)
{\color{black} \citet{dai2015spike} are able to develop a scalable variational inference algorithm for selecting latent inputs in the GP latent variable model (GP-LVM) using the PMF approximation of \citet{carbonetto2012scalable}. By using the sparse inducing points strategy of \citet{titsias2009variational} they are able to recover a closed form variational lower bound to optimise with $\mc O(nm^2)$ complexity. Unfortunately, the same strategy can't be used in the supervised setting when sparsifying the inverse lengthscales, as the expected inverse of the inducing variable gram matrix $\langle K_{ZZ}^{-1}\rangle_{q(\theta, \gamma)}$ is intractable.  \citet{titsias2011spike} also develop an efficient variational EM algorithm for GPs with spike and slab priors in the multi-task learning setting. However their algorithm finds sparse linear mixtures of GPs and so can take advantage of linear model tractabilities.}

By contrast to previous work, our method runs faster than stochastic variational GPR \cite{hensman2013gaussian} up to $n=10^6$, has typically $\mc O(n\log(n)d)$ complexity, and works with any differentiable kernel.

\section{THE SPIKE AND SLAB VARIATIONAL GAUSSIAN PROCESS (SSVGP)}

In this section we {\color{black} derive and present our method in detail. We first present our generative model and variational inference strategy used to overcome the aforementioned intractabilities, as well as our approach for scaling to large-$n$ datasets. We then address our algorithm's sensitivity to hyperparameters and derive a Bayesian Model averaging (BMA) procedure to adapt them reliably to the degree of input relevance sparsity}. In the process we speed up our algorithm substantially through several modifications. We demonstrate performance throughout using a toy example. For key equation derivations see the appendix.

\subsection{Model and Inference Algorithm}

{\color{black} We focus on the standard GPR setting with a
%$p_{\bm \phi}(\bm y|\bm \theta) = \mc N(\bm y|\bm 0,  \tilde K_{XX})$, where
%with zero mean GP priors (i.e. Gaussian noise model with closed form marginal likelihood)  
Gaussian noise model $\mc N(y|f,\sigma^2)$ and zero-mean GP prior $\mc {GP}(f|0,k(\bm x, \bm x'))$, giving rise to closed form marginal likelihood. We place a Gaussian spike and slab prior on the kernel inverse lengthscales.
In doing so, we choose to approximate the Dirac spike with a Gaussian with mass concentrated near zero, which is a strategy that has been previously been used in linear models to improve tractability \cite{george1997approaches, rovckova2014emvs,bai2021spike}.} We also place a Beta prior over the prior inclusion probability. For $n$ samples $(y_i,\bm x_i)_{i=1}^n$, this results in the model%

\vspace{-2em}
{\small{\begin{align*}
p_{\bm \phi}(\bm y|\bm \theta) & = \mc N(\bm y|\bm 0,  \tilde K_{XX})\\
p( \theta_j,\gamma_j|\pi) & = \left[\mc N\left(\theta_j| 0,\frac{1}{cv}\right)\pi\right]^{\gamma_j}\left[\mc N\left(\theta_j| 0,\frac{1}{v}\right)(1-\pi)\right]^{1-\gamma_j} \\
p(\pi) & = \mc Beta(\pi|a,b).
%\label{gen3}
\end{align*}}}%
Here $\bm \theta \in \mathbb R^d$ are the inverse lengthscales\footnote{{\color{black} Note that since most stationary kernel functions are a function of the square of $\theta$ (i.e. $k_{\theta}(x,x') = k_{-\theta}(x,x')$), we let the inverse lengthscales be unconstrained in domain, which enables a Gaussian spike and slab to be used. For any kernel which violates this we place the prior on $\sqrt{\theta}$.}}, 
$\bm \phi = \{\tau, \sigma^2\} \in \mathbb R^2_{>0}$ are the scale and noise parameters, 
$\bm \gamma \in \{0,1\}^d$ are the binary inclusion indicators, $[\tilde K_{XX}]_{ij} =  k_{\tau, \bm \theta}(\bm x_i, \bm x_j)+\delta_{1}(i=j)\sigma^2$, $v\gg1$ is the precision of the spike distribution, and $cv\ll1$ is the precision of the slab distribution.
 
{\color{black} The Gaussian spike approximation enables us to derive a fast approximate co-ordinate ascent variational inference (CAVI) algorithm under the minimal factorisation: $p_{\bm \phi}(\bm \theta, \bm \gamma, \pi|\bm y)\approx q(\bm \theta)q(\bm \gamma)q(\pi)$}, since now the KL term of the evidence lower bound 
{\color{black}
{{\begin{align}\mc F & = \langle \log p(\bm y|\bm \theta) \rangle_{q(\bm \theta)}-KL[q(\bm \theta)q(\bm \gamma)||p(\bm \theta, \bm \gamma)]\label{elbo}\end{align}}}}is well defined. In particular, if $q(\bm \theta)$ is chosen such that $\langle \theta_j^2 \rangle_{q(\theta)}$ has closed form, exact co-ordinate ascent updates for $q(\bm \gamma)$, $q(\pi)$ are available
    \begin{align}
     q^*(\gamma_j) & = \mc Bern(\gamma_j|\lambda_j) \nonumber \\ 
    \lambda_j = &\left(1+c^{-\frac{1}{2}}e^{-\frac{v}{2}\langle \theta_j^2 \rangle_{q(\theta)}(1-c)+\left \langle \log \frac{1-\pi}{\pi} \right \rangle_{q(\pi)}}\right)^{-1} \label{qgamma}\\
    q^{*}(\pi) & = \mc Beta(\pi|\xi_{a},\xi_{b}) \nonumber \\
    (\xi_{a}, \xi_{b}) & = (a+\sum\nolimits_{j=1}^d\lambda_j, b+d-\sum\nolimits_{j=1}^d\lambda_j))\label{qpi}\end{align}

Note $\lambda_j = q(\gamma_j = 1)$ is the PIP for dimension $j$. Whilst exact CAVI updates are not available for $q(\bm \theta)$ due to the non-conjugacy of the model, for a suitable parametric restriction $q_{\bm \psi}(\bm \theta)$ we can approximate its co-ordinate update using stochastic gradient-based optimisation of $\mc F$ with the reparameterisation trick \cite{kingma2013auto}. {\color{black}The reparameterisation trick enables us to unbiasedly approximate the intractable gradient of the expected likelihood term $\nabla_{\bm \psi} \langle \log p(\bm y|\bm \theta) \rangle_{q_{\bm \psi}(\bm \theta)}$} by parameterising $q_{\bm \psi}(\cdot)$  such that $\bm \theta = \mc T_{\bm \psi}(\bm \epsilon)$, where  $\mc T_{\bm \psi}(\cdot)$ is differentiable, $\bm \epsilon \sim q(\bm \epsilon)$ and $q(\bm \epsilon)$ does not depend on $\bm \psi$. {\color{black} We also choose $q_{\bm \psi}(\bm \theta)$ to ensure the KL term has closed form gradients.} In this case, given $S$ samples $\bm \epsilon^{(1:S)} \sim q(\bm \epsilon)$, ${\nabla}_{\bm \psi}  \mc F$ can be unbiasedly approximated as
\begin{align}\hat{\nabla}_{\bm \psi}  \mc F & = \frac{1}{S}\sum \nolimits_{s=1}^S\bigl({\nabla}_{\bm \theta}\log p_{\bm \phi}(\bm y|\bm \theta^{(s)}) \nabla_{\bm \psi} \mc T_{\bm \psi}(\bm \epsilon^{(s)}) \bigr) \nonumber\\ 
& \quad \quad \quad \quad -\nabla_{\bm \psi}KL[q(\bm \theta)q(\bm \gamma)||p(\bm \theta| \bm \gamma)] \label{psi}
\end{align} 

Scale and noise parameters $\bm \phi$ can be optimised jointly using a similar unbiased approximation
\begin{align}\hat{\nabla}_{\bm \phi}  \mc F & = \frac{1}{S}\sum\nolimits_{s=1}^S{\nabla}_{\bm \phi}\log p_{\bm \phi}(\bm y|\bm \theta^{(s)}). \label{alpha}\end{align}
Thus, our approximate CAVI algorithm (a-CAVI) iterates between 1) \emph{exactly} maximising $\mc F$ with respect to $\{q(\bm \gamma)$, $q(\pi)\}$ using Eqns.~\eqref{qgamma} and \eqref{qpi}, and 2) \emph{approximately} maximising $\mc F$ with respect to $\{q_{\bm \psi}(\bm \theta)$, $\bm \phi\}$ using gradient based optimisation with Eqns.~\eqref{psi}, \eqref{alpha}, and the overall parameter update 
%\eqref{grad},
\begin{align}(\bm \psi, \bm \phi) \leftarrow (\bm \psi, \bm \phi) + \bm \eta \odot (\hat{\nabla}_{\bm \psi}  \mc F ,\hat{\nabla}_{\bm \phi}  \mc F).\label{grad}\end{align}
Algorithm 1 summarises the steps. In practice we set learning rates $\bm \eta$ using ADAM \cite{kingma2014adam} and set $q(\bm \theta)$ as a Mean Field Gaussian (MFG) {\color{black} to ensure $\mc O(d)$ time and storage complexity}. In this case $q(\bm \epsilon) = \mc N(\bm 0, I)$ and $\mc T_{\bm \psi}(\bm \epsilon) = \bm \sigma \odot \bm \epsilon + \bm \mu$. Fortunately reparameterisation gradient variance is sufficiently low such that setting $S=1$ enables efficient learning. 
\begin{algorithm}[t]
\SetKwInOut{Input}{Input}
\SetKwInOut{Output}{Output}
{{\Input{ Data: ($\bm y,X$), initialisation: $(\bm \psi^{(0)}, \bm \lambda^{(0)}, \bm \xi^{(0)},\bm \phi^{(0)})$, learning rates:  $\bm \eta$, hyperparameters: ($v,c,a,b$), $\#$ a-CAVI iters: $K$, \# gradient iters: $T$, \# MC samples: $S$}
\For{$K$ iterations} {
 {\For {$T$ steps} {
 
   \textbf{sample $\bm \epsilon^{(1:S)} \sim \mc N(\bm 0, I)$}. \\
\textbf{get $\hat{\nabla}_{\bm \psi}  \mc F$ ,$\hat{\nabla}_{\bm \phi}  \mc F$} using eq. \eqref{psi} and \eqref{alpha}. \\
\textbf{update $(\bm \psi,\bm \phi)$} using equation \eqref{grad}. \\
}
\textbf{update $\bm \lambda$} using equation \eqref{qgamma}. \\
\textbf{update $\bm \xi$} using equation \eqref{qpi}.
}}
\Output{$\bm \psi, \bm \lambda, \bm \xi,\bm \phi$}}}
\caption{a-CAVI for SSVGP training}
\end{algorithm} 

\textbf{Complexity Analysis}: When $q(\bm \theta)$ is MFG and $S=1$, gradients \eqref{psi} and \eqref{alpha} cost $\mc O(n^3)$ for inverting $K_{XX}$ and $\mc O((n^2+1)d)$ for matrix-vector products, which is similar complexity to ML-II gradients. The remaining CAVI updates for $q(\bm \gamma)$, $q(\pi)$ are $\mc O(d)$ and generally extremely cheap to perform. If we had instead used the Dirac spike and PMF approximation as in \citet{dai2015spike}, then no exact CAVI updates would be tractable and only score gradients (i.e. BBVI \cite{ranganath2014black}) would be available to optimise $q(\bm \gamma)$ (without relying on continuous relaxations), which typically require $S\gg 1$ samples per iteration for efficient learning \cite{mohamed2020monte}. Table \ref{t1} demonstrates our method significantly outperforms using a Dirac spike with PMF+BBVI on a toy example, whilst retaining much faster runtimes.

\textbf{Predictions}: At test time, given $S_*$ samples $\bm \theta^{(1:S_*)}\sim q_{\bm \psi}(\bm \theta)$ and learned $\bm \phi$, the predictive distribution can be approximated by a discrete mixture of standard GPR posterior predictives $\{p_{\bm \phi}(y_*|\mc D, \bm x_*, \bm \theta^{(s)})\}_{s'=1}^{S_*}$, which has closed form moments (see the appendix)
\begin{align}p(y_*|\mc D, \bm x_*) \approx \frac{1}{S}\sum\nolimits_{s'=1}^{S_*} p_{\bm \phi}(y_*|\mc D, \bm x_*, \bm \theta^{(s')}) \label{pred}\end{align}

\textbf{Interpreting the Posterior Factorisation}: Our factorisation over inverse lengthscales $\bm \theta$ and inclusion variables $\bm \gamma$ mean that $q(\bm \gamma)$ is not used at test time. Whilst this may seem counter-intuitive, note that $q(\bm \theta)$ automatically embeds information from $q(\bm \gamma)$ during a-CAVI iterations in `message passing' fashion. In the MFG case, the KL term acts as an L2-regulariser on $\bm \psi = \{\bm \mu, \bm \sigma\}$, where the penalty scale varies with $\bm \lambda$. {\color{black} This can be seen by examining the parts of the KL term that depend on variational parameters $\{\mu_j,\sigma_j\}$}
\begin{align}KL_{\mu_j} & = -0.5(\lambda_j{vc}\mu_j^2 + (1-\lambda_j){v}\mu_j^2)\\
KL_{\sigma_j} & =  -0.5(\lambda_j{vc}\sigma_j^2 + (1-\lambda_j){v}\sigma_j^2) + H[q]\end{align}
As a result, when $\lambda_j \approx 0$, we often have $\bm \mu, \bm \sigma \to 0$ for large enough spike precision $v$. Thus, our algorithm can be interpreted as adaptively regularising the posterior over inverse lengthscales based on learned PIPs.

\subsection{Scaling to Large-$n$ Datasets}
Since our algorithm suffers $\mc O(n^3)$ complexity, in large-$n$ datasets we use the approach taken by \citet{chen2020stochastic} of approximating $\nabla\log p_{\bm \phi}(\bm y|\bm \theta)$ at each iteration using a rescaled $m$-sized minibatch $\bm y_{m}$ comprised of a uniformly sampled point $y_i$ and its $m-1$ nearest neighbours $\bm y_{m,\neg i}: \frac{n}{m}\nabla \log p_{\bm \phi}(\bm y_{m}|\bm \theta)$.
%\footnote{In App. 2 we demonstrate this rescaling factor is reasonably accurate even when $n\gg m$}
This reduces the inversion complexity bottleneck to $\mc O(m^3)$ at the expense of an $\mc O(nd)$ nearest neighbour search. Although minibatch gradients are biased due to the non-decomposable likelihood, \citet{chen2020stochastic} were able to outperform a range of state-of-the-art scalable GP approximations on various datasets, and their approach is easily integrated into our base algorithm. When the kernel is monotonically decreasing in a distance metric, we use this metric to find the neighbours with the posterior means $\langle \bm \theta \rangle_{q(\bm \theta)} = \bm \mu$ in place of inverse lengthscales $\bm \theta$. For example using kernels of the form given in eq. \eqref{kernel} we use distance metric $d(\bm x, \bm x') = ||\bm \mu \odot (\bm x - \bm x')||_2$.

Since making predictions also suffers an $\mc O(n^3)$ inversion cost, in $n\gg10^4$ datasets we follow \citet{jankowiak2021scalable} and truncate the predictive distribution in eq. \eqref{pred} using $m_*$ nearest neighbours of $\bm x_{*}$, per each sampled $\bm \theta^{(s)}$. We again use the distance metric in the kernel but replace $\bm \mu$ with $\bm \theta^{(s)}$. This reduces inversion complexity to $\mc O(S_*m_*^3)$ per test point, again at the expense of an $\mc O(S_*nd)$ nearest neighbour search. 

\begin{figure}
\centering
    %\makebox[0.5\textwidth][l]{\includegraphics[width = 0.24\textwidth]{ML_II_q=5_d=100_n=300.png}
    %\includegraphics[width = 0.24\textwidth]{SSVGP_PIPs_q=5_d=100_n=300_v=1.png}}
   \includegraphics[width = 0.48\textwidth]{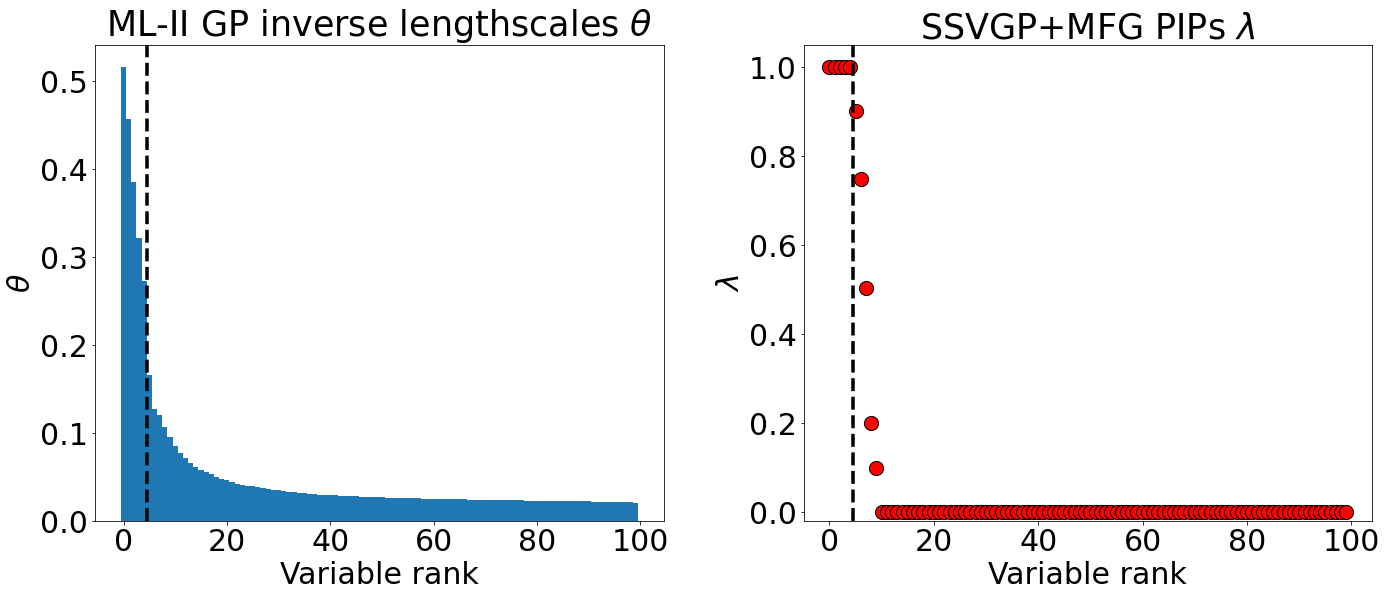}
    \caption{{{Toy example size ordered profile for ML-II GP inverse lengthscales (LHS) and SSVGP+MFG PIPs (RHS). First 5 variables (LHS of black dashed line) are constrained to be the 5 relevant variables. Profiles are averaged over 10 trials. Whilst the ML-II GP profile fails to shrink many irrelevant inverse lengthscales, the SSVGP learns to exclude nearly all irrelevant variables with high probability and includes all relevant variables with near certainty.}}}
    \label{f1}
\end{figure}
\begin{table}[t]
    \centering
    \resizebox{1\columnwidth}{!}{
    \begin{tabular}{ccc}  \hline
    Method     & MSE & Runtime (s) \\ \hline
 ML-II GP      & 0.096 ± 0.014 & \textbf{5.0} ± 0.7   \\
 PMF+BBVI(10)  & 0.088 ± 0.019 & 71.6 ± 1.6  \\
 PMF+BBVI(100) & 0.077 ± 0.013 & 595.3 ± 4.9 \\
 SSVGP+MFG         & \textbf{0.068} ± 0.011 & 27.8 ± 0.5  \\ 
 \hline
    \end{tabular}}
    \caption{{{Toy example mean $\pm$ sd results (10 trials) for ML-II GP, Dirac spike and slab GP with PMF+BBVI using 10 and 100 score-grad samples respectively, and SSVGP with Mean Field Gaussian $q(\bm \theta)$.}}}
    \label{t1}
\end{table}

\subsection{Toy Example Demonstration}

We now demonstrate our SSVGP on a toy example where we draw $n=300$ samples of 100-dimensional input $\bm x \sim \mc N(0,I)$ and set response using 5/100 inputs as $y = \sum_{j=1}^5sin(a_jx_j)+\epsilon$, where $\{a_j\}$ are grid spaced over $[0.5,1]$, and the noise to signal ratio is set to 0.05. We set the spike precision to $v = 10^4$, slab precision to $cv = 10^{-4}$, and Beta prior hyperparameters to $a = b = 10^{-3}$. {\color{black}Note, whilst a $\mc Beta(a,a)$ prior is strongly informative in the tails as $a \to 0^+$, setting $a \approx 0$ closely approximates the improper Haldane prior $\mc Beta(0,0)$, which is flat on $\log(\frac{\pi}{1-\pi})$ (the
statistic used in the a-CAVI algorithm) \cite{zhu2004counter}. We emphasise the Beta hyperprior has negligible effect on our algorithm results, as demonstrated in section 4.} For comparison we implement an ML-II GP using GPyTorch \cite{gardner2018gpytorch} and a spike and slab GP using the Dirac spike and PMF approximation trained using BBVI. See the appendix for implementation and experiment details. In all cases the SE kernel is used. 

 Table \ref{t1} displays (normalised) mean squared error (MSE) and runtime (mean $\pm$ sd) obtained on 10 trials using $n_* = 100$ test points. Our SSVGP outperforms both ML-II and PMF-BBVI on average, and is much closer in runtime to the ML-II GP. In Fig.~\ref{f1} we also highlight how our SSVGP enables much more reliable variable selection than the ML-II GP by better distinguishing between irrelevant and relevant variables.

\section{BAYESIAN MODEL AVERAGING FOR ADAPTIVE SPARSITY}

One problem with our method is that the learned model sparsity (via PIPs) is heavily dependent on spike and slab parameter $v$. This can be seen by analysing the `posterior point of intersection' (PPI), which is the value of transformed expected sufficient statistic $\hat \theta := \sqrt{\langle \theta^2 \rangle_{q(\theta)}}$ that induces an even chance of inclusion $\lambda=\frac{1}{2}$ in the a-CAVI update for $q(\bm \gamma)$
\begin{align}\hat \theta = \sqrt{\frac{\log\left(\frac{1}{c}\right)+ 2\langle \log \left(\frac{1-\pi}{\pi}\right) \rangle_{q(\pi)}}{v(1-c)}}\label{PPI}
\end{align}

The PPI adapts with $q(\pi)$ during training, however the adaptability is limited as the effect is through the square root of the log-odds. {\color{black} For example if a fixed value of $\pi$ is moved from $0.001$ to $0.999$, the PPI only moves from $0.021$ to $0.057$ for $(v,c) = (10^4,10^{-8})$}. Thus, since the PPI is $\mc O(v^{-\frac{1}{2}})$, $v$ can entirely determine the sparsity of the PIPs. This is demonstrated in Table \ref{t2}, which shows that when we run our SSVGP on the same 10 trials increasing $v$ from  $10^2$ to $10^6$, the average recovered PIP increases from 0 to 1. We therefore need a reliable way of adapting $v$ to the input sparsity in the data.

%Whilst $v=10^4$ performed excellently on this problem, this does not hold in general unfornately. For example, when we re-run the SSVGP using $v=10^4$ on 10 trials of a modified version of the design where now $q=40$ signals are relevant and the generating coefficients are now $10\times$ smaller, the learned PIPs are too low for many relevant variables (see figure \ref{f2} below). This is because the generating process is now denser (encouraging a higher PPI through $q(\pi)$) and the optimal inverse lengthscales are much smaller (requiring a smaller PPI for accurate variable selection). 
%\begin{figure}[H]
%    \makebox[0.55\textwidth][l]{\includegraphics[width = 0.25\textwidth]{SSVGP_PIPs_q=5_d=100_n=300_v=4.png}
%    \includegraphics[width = 0.25\textwidth]{SSVGP_PIPs_q=40_d=100_n=300_v=4.png}}     
%    \caption{{{Toy example results (10 trials): average profile of PIPs for SSVGP ($v=10^4$) with Mean Field posterior on (LHS) sparse design with $q=5$ signals and (RHS) dense design with $q=40$ signals.}}}
%    \label{f2}
%\end{figure}
We found directly maximising $\mc F$ with respect to $v$ (or variational posterior $q(v)$) lead to highly initialisation dependent results, and so did not address the problem. Our solution is to instead train a set of models $\{\mc M_k\}_{k=1}^K$ with different scale parameters $v_k \in \mc V$, and Bayesian model average over the learned posteriors  $\{q_k(\bm \theta)_{k=1}^K\},\{q_k(\bm \gamma)_{k=1}^K\}$ using approximate posterior model probabilities $w_k \approx p(\mc M_k|\bm y)$. Marginal PIPs $\{\bar \lambda_j \}_{j=1}^d =  \{\sum_{k=1}^K\lambda^{(k)}w_k\}_{j=1}^d$ can then be used to do {\color{black} variable selection as in the original algorithm}.
\begin{table}
    \centering
%\resizebox{1\columnwidth}{!}{
\begin{tabular}{l|ccccc}
 \hline
$v$         & $10^2$     & $10^3$ & $10^4$ & $10^5$ & $10^6$  \\ \hline
 average PIP   & 0 & 0.05 & 0.07 & 0.34 & 1 \\
 \hline
\end{tabular}
\caption{{{Toy example average PIP recovered for SSVGP with varying spike and slab parameter $v$.}}}
    \label{t2}
\end{table}

However, for this procedure to work we need (1) a fast enough algorithm to enable multiple ($\geq10$) trials without compromising speed and scalability, and (2) a good approximation to the posterior model probability $p(\mc M_k|\bm y)$. We address each of these aspects below.
\subsection{Speeding Up a-CAVI}

We first modify our a-CAVI algorithm to enable order of magnitude speed-ups, whilst only negligibly impacting performance when $v$ is appropriately set.

\textbf{Zero Temperature Restrictions}: We zero temperature (ZT) restrict the inverse lengthscale posterior: $q(\bm \theta) = \delta_{\bm \mu}(\bm \theta)$. This eliminates the need for Monte-Carlo sampling at training and test time as both the expected likelihood term $\langle p_{\bm \phi}(\bm y|\bm \theta)\rangle_{q(\bm \theta)}$ and predictive distribution now have closed form at $\bm \theta = \bm \mu$.

\textbf{Dropout Pruning}: We prune variables with small PIPs during training iterations. This type of procedure has been used to great effect in sparsifying neural networks whilst only negligibly impacting performance \cite{hoefler2021sparsity}. Specifically after every a-CAVI iteration, we permanently enforce $\langle \theta_j \rangle = \mu_j = 0$ if $\lambda_j \leq \epsilon$ for some chosen $\epsilon \in (0,1)$. This reduces training complexity from $\mc O(d)$ to $\mc O(p_t)$, where $p_t = \{\# \lambda^{(t)}_j > \epsilon\}$, leading to  dramatic speed ups when few inputs are relevant. We also prove in the appendix that under certain conditions pruning will only negligibly affect converged solutions of a-CAVI. In practice we set $\epsilon = 0.5$ to ensure non-pruned variables are those with at least an even probability of inclusion, {\color{black}although our algorithm is insensitive\footnote{{\color{black}This is because the PIP update takes the form of generalised logistic funtion $\lambda(\theta) = (1+Ae^{-B\theta^2})^{-1}$ with large positive constants $A,B$. Thus typically $\lambda^* \to \{0 \cup 1\}$.}} to the pruning threshold.}

We also use $m < n$ nearest neighbour minibatching to further improve runtime even when $n<10^3$, as we find for large enough $m$ (i.e. $m \geq \frac{n}{4}$) this does not harm and in some cases improves performance. We believe this is because if $m$ is sufficiently large, the gradient bias remains small enough for the intrinsic regularisation benefits of SGD \cite{smith2021origin} to be leveraged with minimal cost. Table \ref{t_speed} demonstrates the obtainable speed ups using these modifications on the toy example, with pruning threshold $\epsilon = 0.5$ and $m=\frac{n}{4}$ minibatching.  Whilst average performance remains similar, average runtime drops to $1.3s$.

\begin{table}
    \centering
\resizebox{1\columnwidth}{!}{
\begin{tabular}{lllll}
 \hline
                  & MF           & ZT & ZT+drop & ZT+drop(m=n/4)     \\
\hline
 MSE          & 0.068  & 0.068 & 0.064 & 0.065 \\
 avg PIP     & 0.07 & 0.05 & 0.06 & 0.05 \\
 Runtime(s)       & 27.8 & 24.1  & 10.5 & 1.3  \\

 \hline
\end{tabular}}
    \caption{{{Toy example average results for SSVGP implementing a-CAVI modifications sequentially (left to right). Using all modifications speeds up runtime by $\sim 20 \times$, whilst performance remains similar.}}}
    \label{t_speed}
\end{table}

%\begin{figure*}
%\centering
%    \makebox[1\textwidth][c]{\includegraphics[width = 0.33\textwidth]{ML_II_q=5_d=100_n=300.png}
%   \includegraphics[width = 0.33\textwidth]{SSVGP_postmean_q=5_d=100_n=300.png}
%    \includegraphics[width = 0.33\textwidth]{SSVGP_q=5_d=100_n=300.png}}    
%    \label{f1}
%    \caption{{{Toy example results (average size ordered variable rankings over 10 trials): Left =  vanilla (ML-II) GP inverse lengthscale profile, Middle = SSVGP posterior mean inverse lengthscale profile $\langle \bm \theta \rangle_{q(\bm \theta)}$, Right = SSVGP posterior inclusion probabilities (PIPs) profile. LHS of black dashed line constrained to be 5 relevant variables. Whilst the vanilla GP fails to shrink the inverse lengthscales sufficiently for the 95 irrelevant variables and displays a smooth gradual decline in the profile, our SSVGP shrinks the irrelevant posterior mean inverse lengthscales to near zero values ($\ll10^{-3}$) and learns to exclude almost all irrelevant variables with high probability.}}}
%\end{figure*}

\subsection{Approximating Posterior Weights with the Leave-One-Out Predictive Density}
Rather than use the evidence lower bound $\mc F$ to approximate the posterior model weights (under a uniform prior over models $e^{F_k} \leq p(\bm y|\mc M_k) \propto  p(\mc M_k|\bm y)$), we use the leave-one-out predictive density (LOOPD): $\prod_{i=1}^np(y_i|\bm y_{\neg i}, \mc M_k)$ as in \citet{yao2018using}. We do not use $\mc F$ as its preference for model sparsity is heavily influenced by hyperparameter $c${\color{black}, which controls the KL divergence between the spike and slab of the prior}. This arises due to the fixed variable inclusion cost of $\mc O(\log(\frac{1}{c}))$ {\color{black}from the normaliser of the slab distribution}. Additionally, the LOOPD known to be more robust against overfitting and model mis-specification than the evidence $p(\bm y|\mc M_k)$ itself \cite{jankowiak2021scalable, rasmussen2003gaussian}.

Under the ZT posterior, inverse lengthscales $\bm \theta$ can be treated as parameters fixed at $\bm \mu_k$ for model $\mc M_k$. In this case the LOOPD for $\mc M_k$ is that of a standard GPR evaluated at ($\bm \theta = \bm \mu_k, \bm \phi = \bm \phi_k)$. To compute the LOOPD in $\mc O(n^3)$ time in small datasets we use the \citet{burkner2021efficient}
%\citeauthor{burkner2021efficient}'s \citeyear{burkner2021efficient}
%Burkner et al's (2020) 
algorithm, which uses Rank-1 updates to $K_{XX}$. In large datasets we nearest-neighbour truncate the LOOPD following the success of this approach in \citet{jankowiak2021scalable}. That is, for each ($y_i,\bm x_i$), we condition on $\tilde m$ nearest neighbours of $\bm x_i$ found using the distance metric in the kernel. This reduces complexity to $\mc O(m^3n)$ but introduces an $\mc O(n f(n)p_k)$ query cost, where $p_k$ is the count of non-pruned variables in $\mc M_k$ and $f(n) \in [\log(n),n]$ is the query cost using Ball-trees \cite{omohundro1989five} or KD-trees \cite{bentley1975multidimensional} (our preferred algorithms).

Given the recovered posteriors and model weights, the predictive distribution is a discrete mixture of standard GP predictive posteriors evaluated at $\{\bm \mu_k, \bm \phi_k\},$
%
%\vspace{-1em}
{%\small
{\begin{align}
p(\bm y_*|X_*, \mc D) & \approx \sum\nolimits_{k=1}^K\int p_{\bm \phi_k}(\bm y_*|X_*, \bm \theta, \mc D)q_k(\bm \theta)d\bm \theta w_k \nonumber \\
& = \sum\nolimits_{k=1}^K p_{\bm \phi_k}(\bm y_*|X_*, \bm \theta=\bm \mu_k, \mc D)w_k. \label{pred0}
\end{align}}}To speed up prediction time we stochastically threshold the {\color{black}recovered posterior model weights $\bm w$ by updating $\bm w \to \tilde {\bm w} = \frac{\bm z}{S}$, where $\bm z$ is drawn from a Multinomial distribution using probability vector $\bm w$ and $S$ trials. Thus we have that $\mathbb E[\tilde {\bm w}] = \bm w$ and  $p(\tilde w_i = 0|w_i = \frac{1}{S}) = \left(\frac{S-1}{S}\right)^S \approx \frac{1}{3}$ for $S \geq 100$}. Additionally, when predictive accuracy is the core focus we simply select the best model. We use nearest-neighbour truncation at test time for large-$n$ datasets (e.g. $n>10^4$).

\begin{figure}
\centering
    %\makebox[0.5\textwidth][l]{\includegraphics[width = 0.24\textwidth]{SSVGP_postmean_q=5_d=100_n=300.png}
    %\includegraphics[width = 0.24\textwidth]{SSVGP_q=5_d=100_n=300.png}}  
    \includegraphics[width = 0.48\textwidth]{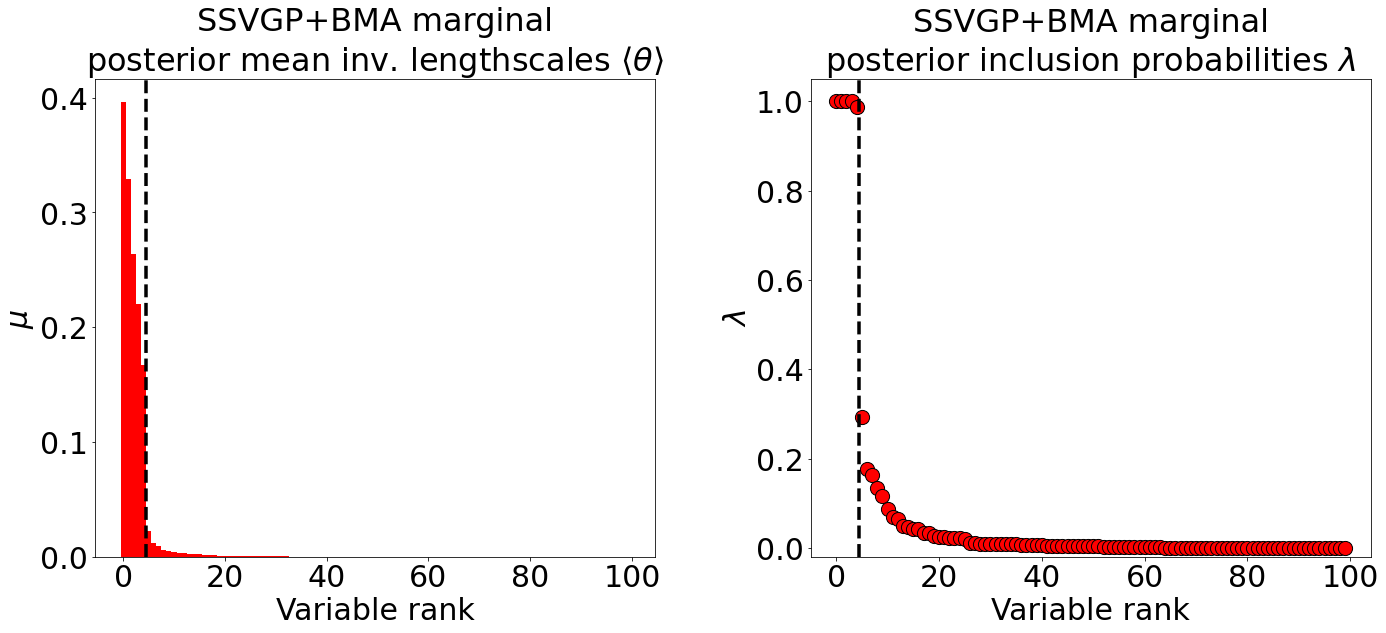}  
    \caption{{{Toy example average size ordered profiles for (LHS) SSVGP+BMA posterior mean inverse lengthscales and (RHS) SSVGP+BMA PIPs. The PIP profile is better than the SSVGP+MFG (Fig.~\ref{f1}) and the posterior mean inverse lengthscales are shrunk to near-zero for all irrelevant dimensions due to the regularisation from $\{q_k(\bm \gamma)\}_{k=1}^K$ and learned model weighting.}}}
    \label{f2}
\begin{table}[H]
\resizebox{1\columnwidth}{!}{
\begin{tabular}{llll}
 \hline
Method       & MSE         & MCC   & Runtime (s)    \\
\hline
 ML-II GP            & 0.096 ± 0.014 & -   & \textbf{5.0} ± 0.7   \\
 PMF+BBVI(100)    & 0.077 ± 0.013 & 0.62 ± 0.06 & 595.3 ± 4.9 \\ 
 SSVGP+MFG & 0.068 ± 0.011 & 0.82 ± 0.09 & 27.8 ± 0.5  \\
  SSVGP+BMA & \textbf{0.064} ± 0.01 & \textbf{0.98} ± 0.04 & 14.6 ± 0.4 \\
\hline
\end{tabular}}
\caption{{{Toy example mean $\pm$ sd results. SSVGP+BMA outperforms all other methods in predictive accuracy (MSE) and variable selection accuracy (MCC), and runs in similar time to ML-II GP.}}}
    \label{t3}
\end{table}
\end{figure}

\subsection{BMA on the Toy Example}
We demonstrate our SSVGP with BMA on the same 10 trials of the toy example using 11 values of $v$ grid spaced over $10^4 \times 2^{\{-log_2(1000),...,log_2(1000)\}}$ such that $10^1 \leq v \leq 10^7$, and use the ZT posterior with $m=\frac{n}{4}$ minibatching and pruning threshold  $\epsilon = 0.5$.

We display in Table \ref{t3} performance results against previous models and in Fig.~\ref{f2} the average profile of PIPs and posterior mean inverse lengthscales. Note we also report the Matthews Correlation Coefficient (MCC $\in [-1,1]$) \cite{matthews1975comparison} to measure variable selection accuracy when using $\bar \lambda = 0.5$ as an inclusion threshold. Our SSVGP with BMA outperformed all other methods in terms of predictive and variable selection accuracy, learned a better quality PIP profile than the original SSVGP+MFG without BMA and also ran faster on average ( $\sim 15s$). Thus, our BMA procedure and a-CAVI modifications induced better and faster performance than using our original algorithm, even with a well calibrated choice of $v$.
%\begin{table}
%    \centering
%\resizebox{1\columnwidth}{!}{
%\begin{tabular}{llll}
%\hline \hline
%Algorithm       & MSE         & MCC   & Runtime (s)  \\
%\hline
%  ML-II GP        & 0.089 ± 0.018 & - & \textbf{4.9} ± 0.2\\
% SSVGP+MF($v=10^4$) & 0.095 ± 0.019 & 0.77 ± 0.1   & 24.2 ± 0.7 \\
% SSVGP+ZTdrop+BMA & \textbf{0.086} ± 0.016 & \textbf{0.82} ± 0.09 & 16.7 ± 0.7  \\\hline \hline
%\end{tabular}}
%\caption{{{Toy example ($q=40$ design) mean $\pm$ sd results for ML-II GP, baseline SSVGP+MF with $v=10^4$, and adaptive SSVGP+BMA.}}}
%    \label{t4}
%\end{table}
\section{EXPERIMENTS}
We now test our SSVGP in several experiments (we use BMA with ZT and dropout by default from now on). We consider a synthetic small-$n$ experiment, a synthetic large-$n$ experiment and benchmark real datasets. See the appendix for experiment and implementation details. The SE kernel is used throughout, but our method works with any differentiable kernel.
\begin{figure}
\centering
    %\makebox[0.5\textwidth][l]{
    \includegraphics[width = 0.48\textwidth]{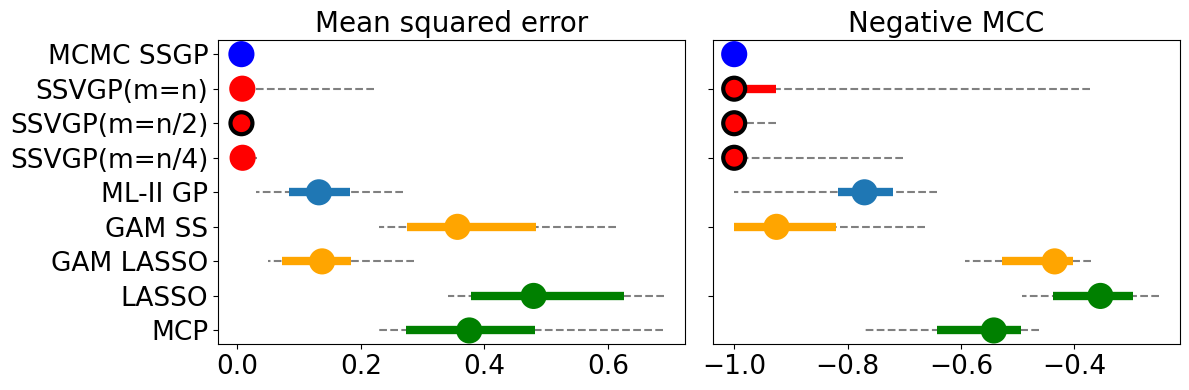}%}
     %\makebox[0.5\textwidth][l]{\includegraphics[width = 0.49\textwidth]{E1_plots_runtime.png}}  
    \caption{{{Experiment 1 median (circle), 25-75 percentiles (line), and 10-90 percentiles (dashed line) results. Black border denotes best median performance. Left is better. SSVGPs outperform all implemented comparators and matched single trial performance of \citet{savitsky2011variable} MCMC spike and slab GP.  }}}
    \label{f3}

\begin{table}[H]
    \centering
\resizebox{1\columnwidth}{!}{
\begin{tabular}{lllll}
 \hline
                        MCMC & SSVGP(n) & SSVGP(n/2) & SSVGP(n/4) & ML-II \\ \hline
   10224s & 20.3s & 11.0s & 8.4s & 3.9s \\
 \hline
\end{tabular}}
\caption{{{Experiment 1 average runtimes for different methods and single trial time of \citet{savitsky2011variable} spike and slab GP when running $5 \times 10^5$ MCMC iters. }}}
    \label{t4}
\end{table}
\end{figure}
\subsection{Experiment 1: A Small-Scale, High Dimensional Variable Selection Problem}
We first test our SSVGP on 50 replications of \citet{savitsky2011variable} main experiment to assess whether our algorithm can compete with their spike and slab GP using a Dirac spike and MCMC. The experiment draws $n=100$ training and $n_*=20$ test samples of $d=1000$ dimensional inputs $\bm x \sim Unif[0,1]^d$ and sets the response as an additive function of 6/1000 inputs: $y = x_1+x_2+x_3+x_4+sin(3x_5)+sin(5x_6)+\epsilon$ for $\epsilon \sim \mc N(0,0.05)$. We also implement the ML-II GP and several benchmark variable selection algorithms: LASSO \cite{tibshirani1996regression}, MCP \cite{zhang2010nearly}, GAM with group LASSO penalty \cite{huang2009group} and GAM with group spike and slab LASSO penalty \cite{bai2020spike}. We implement our SSVGP with minibatch sizes $m\in \{\frac{n}{4},\frac{n}{2}, n\}$ and the same settings as previously. For the ML-II GP we report MCC using the \emph{best} performing threshold from $0.1^{\{0,0.5,1,1.5,2\}}$ as indicative 'best-case' performance with thresholding.

 \begin{figure}
\centering
    %\makebox[0.5\textwidth][c]{
    \includegraphics[width = 0.45\textwidth]{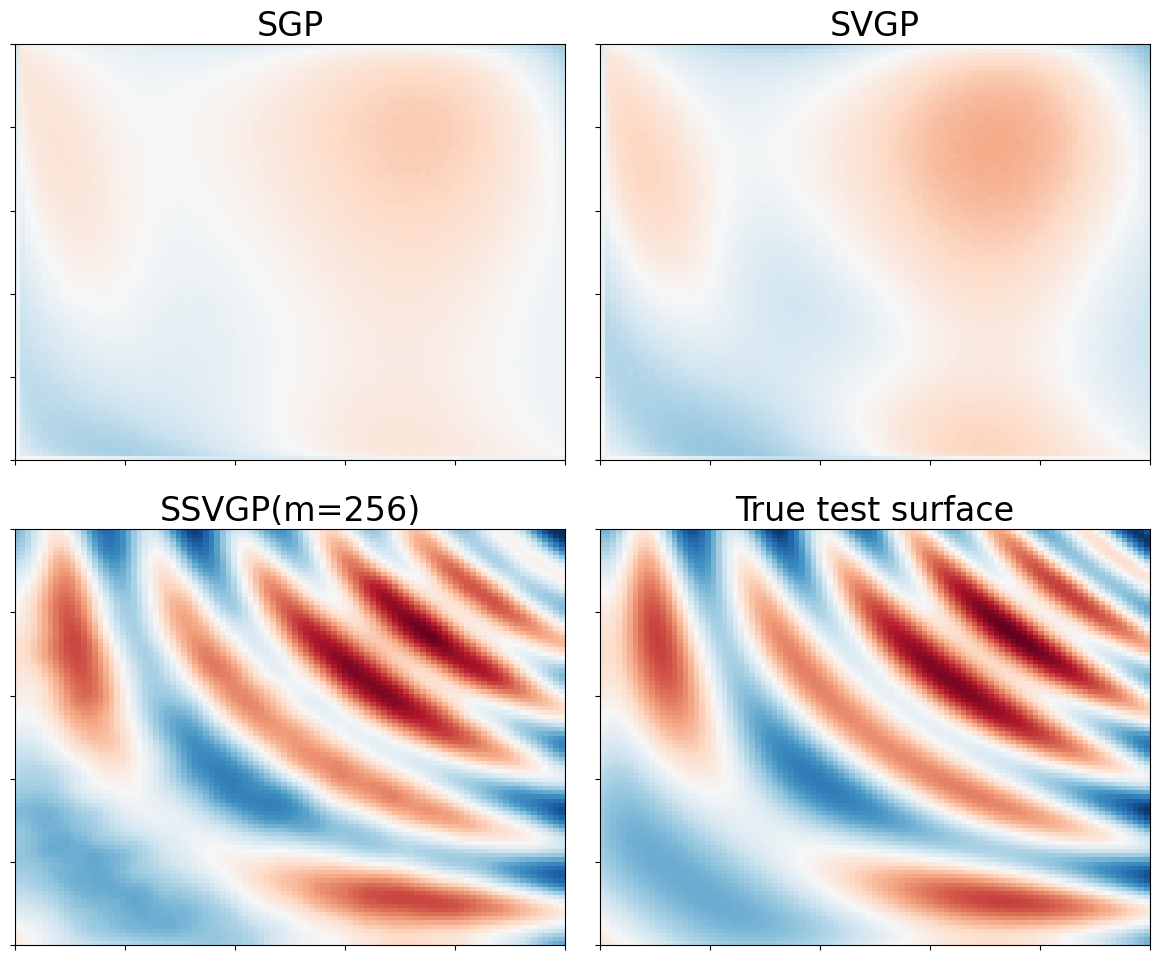}%}  
    \caption{{{Experiment 2 average prediction surfaces as a function of the two relevant inputs when $n=d=10^4$, and true test surface for latent (noiseless) $\bm f_*$. The SGP and SVGP cannot accurately recover the surface due to the number of noise dimensions, whilst our SSVGP recovers a near perfect surface.}}}
    \label{f4}
\end{figure}
In Fig.~\ref{f3}  and Table~\ref{t4} we present MSE,  negative MCC and runtime percentiles for all models against \citet{savitsky2011variable} single completed trial. All SSVGPs outperformed the implemented comparators, and the median trial performance is near identical to the single trial of Savitsky et al's MCMC spike and slab GP but with average runtimes of $<20$ seconds rather than $>10^4$ seconds. When $m=\frac{n}{2}$ performance is best but $m=\frac{n}{4}$ outperformed $m=n$ on average.

\subsection{Experiment 2: A Large Scale Sparse Prediction Challenge}

We next test the scalability and predictive performance of our method in a large-scale synthetic experiment against the Sparse GP (SGP) of \citet{titsias2009variational} and the Stochastic Variational GP (SVGP) of \citet{hensman2013gaussian}, as two popular best-practice benchmarks in the scalable GP literature. We generate synthetic data from a 2d input interactive function $y(\bm x) = \tan(x_1)+\tan(x_2)+\sin(2\pi x_1) +\sin(2\pi x_2)  +\cos(4\pi^2x_1x_2)+\tan(x_1x_2)+\epsilon$ where the noise to signal ratio is set to $\frac{1}{3}$ and $x_1,x_2 \sim \mc U[0,1]$. We then augment with $d-2$ noise dimensions correlated 0.5 with $x_1,x_2$. We fix $nd=10^8$ and run 3 trials of (i) an ultra-high dimensional design $(n=10^4,d=10^4)$, and (ii) a one million training point design $(n=10^6,d=10^2)$. We test on $n_*=10^4$ datapoints but with $x_1,x_2$ grid spaced on $[0,1]$.
\begin{figure}
\centering
    %\makebox[0.5\textwidth][l]{
    \includegraphics[width = 0.48\textwidth]{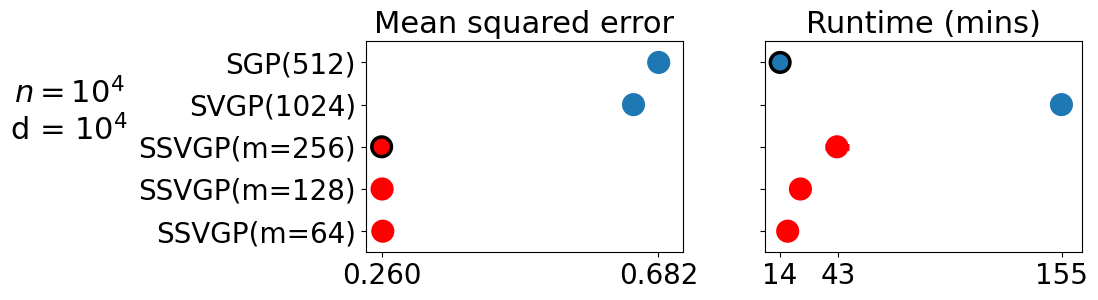}%}
     \\%\makebox[0.5\textwidth][l]{
     \includegraphics[width = 0.48\textwidth]{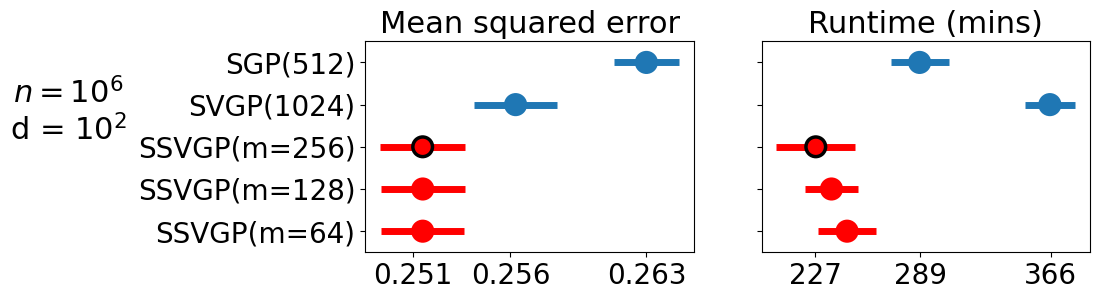}%}  
        \caption{{{Experiment 2 mean (circle) $\pm$ standard deviation (line) results. Black border denotes best on average. Our SSVGPs perform better on average and produce competitive runtimes. Note $m=256$ is fastest when $n=10^6$ as runtime is dominated by LOOPD computation, and since $m=256$ selected fewer variables per model in BMA this reduced compute time.}}}
    \label{f5}
\end{figure}
We implement the SGP and SVGP using GPyTorch \cite{gardner2018gpytorch} with $512$ and $1024$ inducing points respectively consistent with standard practice \cite{chen2020stochastic,wang2019exact,jankowiak2021scalable}. We implement the SSVGP with $m=\{64,128,256\}$ minibatch sizes, and use 64 and 256 neighbours in LOOPD and predictive distribution truncation respectively. In Figure~\ref{f4} we present average prediction surfaces as a function of the two relevant inputs, against the latent test function $\bm f_*$. Figure~\ref{f5} displays MSE and runtime results.
 
 Our SSVGP outperformed the SGP and SVGP on average in both designs and consistently ran faster than the SVGP. Whilst the MSE improvement is minimal in the $d=10^2$ case, when $d=10^4$ the SGP and SVGP are non-competitive. This demonstrates the gains to using our method as input dimensionality and sparsity grows. SSVGP performance is similar across minibatch sizes, but $m=256$ is best as expected.

\begin{figure}
\centering
    \makebox[0.5\textwidth][l]{\includegraphics[width = 0.48\textwidth]{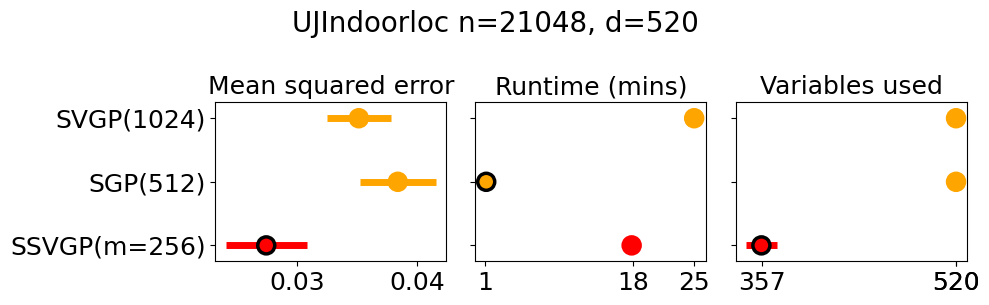}}
     \makebox[0.5\textwidth][l]{\includegraphics[width = 0.48\textwidth]{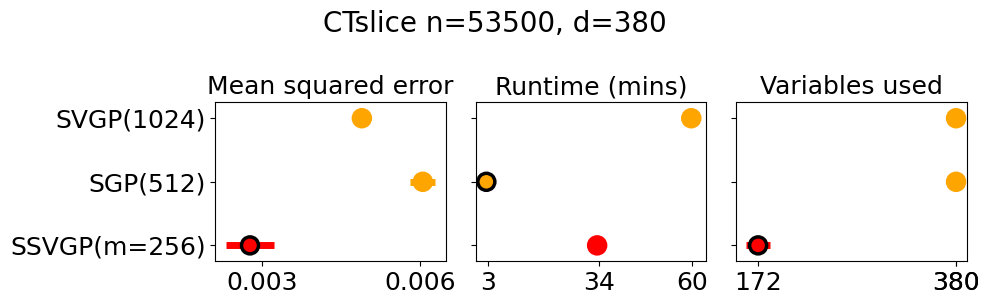}}  
    \caption{{{Real dataset mean (circle) $\pm$ standard deviation (line) results on CTslice and UJindoorloc. Black border denotes best average performance. The SSVGP outperformed both the SGP and SVGP whilst predicting with a sparse subset of identified relevant variables.}}}
    \label{f6}
\end{figure}

\textbf{Computational Cost:} As $n$ grows, SSVGP runtime is dominated by LOOPD computation due to the $\mc O(nf(n)p_k)$  nearest neighbour search cost where $f(n) \in [\log(n),n]$ depending on $p_k$ (\# non-pruned variables in model $k$). As $p_k$ grows, complexity converges to $\mc O(n^2p_k)$ and so in large-$p_k$ scenarios our method may by slower than the SVGP when $n\gg 10^6$. However in scenarios where $p_k \ll 100$ for most of the $K$ models in BMA we expect our method to remain faster than SVGP even when $n\gg 10^6$.
\subsection{UCI Repository Datasets}
We test our method on two high-dimensional datasets from the UCI repository: UJINDOORLOC $(n,d) = (21048,520)$ and CTSLICE $(n,d) = (53500,380)$. We implement the SSVGP with $m=256$ minibatching, and again use 64 and 256 nearest neighbours respectively in LOOPD and predictive distribution truncation. In Figure~\ref{f6} we produce MSE and runtime results from 10 trials of random 4:1 training/test splits, and compare against the the SGP(512) and SVGP(1024). Our SSVGP performed best on average in both datasets and again ran faster than the SVGP. 

\section{CONCLUSION}

 We introduced the spike and slab variational Gaussian process (SSVGP): a fast and scalable training and inference method for Gaussian processes with spike and slab variable selection priors. Unlike previous spike and slab GP implementations for variable selection, our method runs in similar time to sparse variational GPs even on $n=10^6$ sized datasets, has typically $\mc O(n\log(n)d)$ complexity, and works with any differentiable kernel. Thus, our SSVGP captures most of the benefits of spike and slab priors over ARD in high-dimensional designs, without the typical inference cost. Future research could explore Bayesian optimisation \cite{shahriari2015taking} for efficient model search, or extensions to classification and deep kernel learning \cite{wilson2016deep}. {\color{black} For classification the P\'oly{a} Gamma auxilliary variable strategy in \citet{jankowiak2021scalable} could represent a sensible starting point, but we leave this to future work.} 

\clearpage
\nocite{liu2019rao,salimans2014using,bai2021spike,chicco2020advantages,breheny2021package}

\bibliography{references}

\end{document}

\appendix

\newcommand{\mc}{\mathcal}
\runningtitle{Fast and Scalable Spike and Slab Variable Selection in High-Dimensional Gaussian Processes}

\runningauthor{Hugh Dance,Brooks Paige}

% Supplementary material: To improve readability, you must use a single-column format for the supplementary material.
\onecolumn
\aistatstitle{Appendix for `Fast and Scalable Spike and Slab Variable Selection in High-Dimensional Gaussian Processes'}
{
  \hypersetup{linkcolor=black}
  \tableofcontents
}
\section{MATHEMATICAL APPENDIX}

\subsection{Kernel parameterisation}\label{kernel_param}

In this work we treat the inverse lengthscales of the kernel as being unconstrained in their parameterisation: $\bm \theta \in \mathbb R^d$. We note that although it is typical for inverse lengthscales to be strictly positive $\bm \theta \in \mathbb R^d_{>0}$, most kernel functions are invariant to the sign of $\theta_j$ for standard parameterisations. For example any stationary kernel function of the form given in eq. \ref{kernel} in the main text (e.g. squared exponential, Mat\'{e}rn, Cauchy) is  a function $g_{\bm x, \bm x'}(\theta_1^2,...\theta_d^2)$. The linear kernel, polynomial kernel and periodic wrappers on anisotropic stationary kernels are also invariant to $\text{sign}(\theta_j)$. For any kernel where this invariance does not hold our procedure reparameterises the kernel in terms of $\tilde {\bm \theta} = \bm \theta^2$, which restores this property. Thus initial use of a Mean-Field Gaussian posterior approximation $q_{\bm \psi}(\bm \theta)=\prod_{j=1}^d\mc N(\theta_j|\mu_j,\sigma^2_j)$ over the inverse lengthscales is appropriate. Thus, any plots of posterior means $\{\mu_j\}_{j=1}^d$ using our method (the SSVGP) in the main text and this Appendix plot $\{|\mu_j|\}_{j=1}^d$.

\subsection{SSVGP a-CAVI Algorithm}

Here we derive the mathematical form of the reparameterisation gradient approximations of the evidence lower bound (ELBO) $\nabla_{\bm \psi} \mc {\hat F}$, $\nabla_{\bm\alpha} \mc {\hat F}$ and fixed point updates for variational parameters $\bm \lambda, \bm \xi$ and in a-CAVI algorithm 1. We derive these for the case that $q_{\bm \psi}(\bm \theta)$ Mean-Field Gaussian, and discuss simplifications to this when the zero-temperature posterior restriction is used (i.e. in our main method).

\subsubsection{ELBO Gradients}
We first derive expressions \ref{psi} and \ref{alpha} in the main text for  $\nabla_{\bm \psi} \mc {\hat F}$ and  $\nabla_{\bm \phi} \mc {\hat F}$  respectively. We start with expression \ref{elbo} in the main text for the evidence lower bound: 
\begin{align}\mc F= \langle  \log p_{\bm \phi}(\bm y|\bm \theta)\rangle_{q_{\bm \psi}(\bm \theta)} - KL[q_{\bm \psi}(\bm \theta)q(\bm \gamma)q(\pi)||p(\bm \theta, \bm \gamma, \pi)]\end{align}
Taking gradients wrt. $\bm \psi$ we have:
\begin{align}\nabla_{\bm \psi}\mc F= \nabla_{\bm \psi}\langle  \log p_{\bm \phi}(\bm y|\bm \theta)\rangle_{q_{\bm \psi}(\bm \theta)} -  \nabla_{\bm \psi}KL[q_{\bm \psi}(\bm \theta)q(\bm \gamma)q(\pi)||p(\bm \theta, \bm \gamma, \pi)]\label{elbograd}\end{align}
The first term involves the gradient of an expectation of both (i) a log-determinant and (ii) inverse of the gram matrix $\tilde K_{XX} = K_{XX}+\sigma^2I$. This is intractable to compute for most kernel functions, and thus needs to be approximated We use the reparameterisation trick \cite{kingma2013auto} for this. The KL gradient however is exactly tractable. 

For the reparameterisation trick to apply, $q_{\bm \psi}(\bm \theta)$ must be parameterised such that $\bm \theta = \mc T_{\bm \psi}(\bm \epsilon)$ and  $\bm \epsilon \sim q(\bm \epsilon)$. Here $\mc T_{\bm \psi}(\cdot)$ is a transformation differentiable in $(\bm \psi, \bm \epsilon)$, and $q(\bm \epsilon)$ is a base distribution that does not depend on $\bm \psi$. In this case, using the law of the unconscious statistician and the chain rule we have the following: 
\begin{align}
     \nabla_{\bm \psi}\langle  \log p_{\bm \phi}(\bm y|\bm \theta)\rangle_{q_{\bm \psi}(\bm \theta)} & = \nabla_{\bm \psi}\int q_{\bm \psi}(\bm \theta)\log p_{\bm \phi}(\bm y|\bm \theta) d\bm \theta \\
     & = \nabla_{\bm \psi}\int q(\bm \epsilon)\log p_{\bm \phi}(\bm y|\bm \theta)\Big|_{\bm \theta = \mc T_{\bm \psi}(\bm \epsilon)} d\bm \epsilon \\
     & = \int q(\bm \epsilon)\nabla_{\bm \theta}\log p_{\bm \phi}(\bm y|\bm \theta)\Big|_{\bm \theta = \mc T_{\bm \psi}(\bm \epsilon)} \nabla_{\bm \psi}\mc T_{\bm \psi}(\bm \epsilon)d\bm \epsilon  \\
     & = \langle \log p_{\bm \phi}(\bm y|\bm \theta) \nabla_{\bm \psi}\mc T_{\bm \psi}(\bm \epsilon)\rangle_{q(\bm \epsilon)}\label{repgrad}
\end{align}

Thus, given $S$ Monte Carlo samples $\bm \epsilon^{(1:S)}\sim q(\bm \epsilon)$ our unbiased approximation to $\nabla_{\bm \psi} \mc {F}$ is:
\begin{align}\hat{\nabla}_{\bm \psi}\mc F =  \frac{1}{S}\sum_{s=1}^S \left({\nabla}_{\bm \theta}\log p_{\bm \phi}(\bm y|\bm \theta)\Big|_{\bm \theta = \mc T_{\bm \psi}(\bm \epsilon^{(s)})} \nabla_{\bm \psi} \mc T_{\bm \psi}(\bm \epsilon^{(s)})\right)- \nabla_{\bm \psi}KL[q_{\bm \psi}(\bm \theta)q(\bm \gamma)||p(\bm \theta|  \bm \gamma)] \label{g_f}\end{align}
Note here that we have used that $\nabla_{\bm \psi}KL[q_{\bm \psi}(\bm \theta)q(\bm \gamma)q(\pi)||p(\bm \theta,  \bm \gamma,\pi)]  = \nabla_{\bm \psi}KL[q_{\bm \psi}(\bm \theta)q(\bm \gamma)||p(\bm \theta|  \bm \gamma)] $. Note for Mean-Field Gaussian $q_{\bm \psi}(\bm \theta) = \mc N(\bm \theta|\bm \mu,Diag(\sigma_1^2,...,\sigma_d^2))$ we set  $\mc T_{\bm \psi}(\bm \epsilon) = \bm \sigma \odot  \bm \epsilon + \bm \mu$ and $q(\bm \epsilon) = \mc N(\bm 0, I)$.

The equivalent gradient approximation for $\hat \nabla_{\bm \phi}\mc F$ given in the main text is trivial to derive:
\begin{align}\hat \nabla_{\bm \phi}\mc F & = \nabla_{\bm \phi}\langle  \log p_{\bm \phi}(\bm y|\bm \theta)\rangle_{q_{\bm \psi}(\bm \theta)} \\
& = \int \nabla_{\bm \phi}  \log p_{\bm \phi}(\bm y|\bm \theta)q_{\bm \psi}(\bm \theta)d \bm \theta \\
\hat \nabla_{\bm \phi}\mc F & \approx \frac{1}{S} \sum_{s=1}^S \nabla_{\bm \phi}  \log p_{\bm \phi}(\bm y|\bm \theta^{(s)})  :  \bm \theta^{(s)} \sim q_{\bm \psi}(\bm \theta^{(s)})\end{align}

We now derive the closed forms of the main components in the gradient approximations above.  Note that when $\log p_{\bm \phi}(\bm y|\bm \theta) = \log \mc N(\bm y|\bm 0, \tilde K_{XX})$ with $n \times n$ gram matrix $\tilde K_{XX}$ comprised of covariance kernel function evaluations and the additive noise term $[\tilde K_{XX}]_{ij} = k_{\bm \theta, \tau}(\bm x_i, \bm x_j)+ \delta_{0}(i=j)\sigma^2$, then using standard results derived in \citet{rasmussen2003gaussian} we have: 
\begin{align}\nabla_{\theta_j}\log _{\bm \phi}p(\bm y|\bm \theta) = Tr\left[(\bm \alpha \bm \alpha^T - \tilde K_{XX}^{-1})\frac{d\tilde K_{XX}}{d\theta_j}\right] \quad : \quad \bm \alpha = \tilde K_{XX}^{-1} \bm y \label{rasmus}\end{align}
\begin{align}\nabla_{\phi_j}\log _{\bm \phi}p(\bm y|\bm \theta) = Tr\left[(\bm \alpha \bm \alpha^T - \tilde K_{XX}^{-1})\frac{d\tilde K_{XX}}{d \phi_j}\right] \quad : \quad \bm \alpha = \tilde K_{XX}^{-1} \bm y \label{rasmus}\end{align}

Menawhile, the closed form for the gradient of the KL term wrt. $\bm \psi$ is:
\begin{align}\nabla_{\bm \psi}KL[q_{\bm \psi}(\bm \theta)q(\bm \gamma)||p(\bm \theta|  \bm \gamma)]  & = \nabla_{\bm \psi}\int \sum_{\bm \gamma} q_{\bm \psi}(\bm \theta) q(\bm \gamma) \log p(\bm \theta|\bm \gamma)d\bm \theta  + \nabla_{\bm \psi} H[q_{\bm \psi}(\bm \theta)]\\
& = \nabla_{\bm \psi} \sum_{j=1}^d\Big [ \langle \gamma_j \rangle_{q(\gamma_j)}\langle \log \mc N(\theta_j|0, (cv)^{-1})\rangle_{q_{\bm \psi_j}(\theta_j)} \nonumber \\
& \quad \quad \quad \quad +(1-\langle \gamma_j \rangle_{q(\gamma_j)})\langle \log \mc N(\theta_j|0, v^{-1})\rangle_{q_{\bm \psi_j}(\theta_j)} \Big] + \nabla_{\bm \psi} H[q_{\bm \psi}(\bm \theta)]\\
& = -\frac{1}{2}\nabla_{\bm \psi} \sum_{j=1}^d\Big [ \lambda_j cv \langle \theta_j^2 \rangle_{q_{\bm \psi}(\bm \theta)}+(1-\lambda_j)v \langle \theta_j^2 \rangle_{q_{\bm \psi}(\bm \theta)} \Big] + \nabla_{\bm \psi} H[q_{\bm \psi}(\bm \theta)]\\
& = -\frac{v}{2}\nabla_{\bm \psi} \sum_{j=1}^d(\lambda_j c +1-\lambda_j)\langle \theta_j^2 \rangle_{q_{\bm \psi}(\bm \theta)} + \nabla_{\bm \psi} H[q_{\bm \psi}(\bm \theta)]\end{align}
Note that $ H[q_{\bm \psi}(\bm \theta)]$ is the Shannon entropy of $q_{\bm \psi}(\bm \theta)$. Substituting in $q_{\bm \psi}(\bm \theta)  = \prod_{j=1}^d\mc N(\theta_j|\mu_j, \sigma^2_j)$, and using standard Gaussian algebra we get the equations in the main text.

Under the zero-temperature restriction ($q_{\bm \psi}(\bm \theta) = \prod_{j=1}^d \delta_{\mu_j}(\theta_j)$) the gradient approximations are unchanged, except now the approximation is exact for a single sample from the point mass ($\bm \theta^{(s)} =  \bm \mu$), and $\bm \psi = \bm \mu$.

\subsubsection{Fixed Points}
Starting with the standard CAVI update expression in \citet{blei2017variational}, the update for $q(\bm \gamma)$ is given by: 
\begin{align}q(\bm \gamma) & \propto \exp \left\{ \left \langle \log p(\bm \theta, \bm \gamma|\pi) \right \rangle_{q_{\bm \psi}(\bm \theta)q(\pi) } \right\} \nonumber \\
& \propto \prod_{j=1}^d\exp \Big\{ \gamma_j \left \langle \log \mc N(\theta_j|0,(cv)^{-1})\right \rangle_{q_{\bm \psi_j}(\theta_j)}+(1-\gamma_j) \left \langle \log \mc N(\theta_j|0,v^{-1})\right \rangle_{q_{\bm \psi_j}(\theta_j)} \nonumber \\
& \quad \quad +\gamma_j\langle \log \pi \rangle_{q(\pi)}+(1-\gamma_j)\langle \log(1-\pi) \rangle_{q(\pi)}\Big\} \end{align}

Passing over the Gaussian expecations, we have due to the factorisation of the RHS that:
\begin{align}q(\gamma_j=1) & = \lambda_j  \propto c^{\frac{1}{2}}e^{-\frac{1}{2}cv \langle \theta_j^2 \rangle_{q_{\bm \psi_j}(\theta_j)}+ \langle \log \pi \rangle_{q(\pi)}}\\
q(\gamma_j=0) & \propto e^{-\frac{1}{2}v \langle \theta_j^2 \rangle_{q_{\bm \psi_j}(\theta_j)}+ \langle \log (1-\pi) \rangle_{q(\pi)}} \end{align}

After renormalisation we arrive at the update used in a-CAVI algorithm 1:
\begin{align} \lambda_j  & = \left(1+\left(\frac{1}{c}\right)^{\frac{1}{2}}e^{-\frac{1}{2}\langle \theta_j^2 \rangle_{q_{\bm \psi_j}(\theta_j)}v(1-c)+ \langle \log \left(\frac{1-\pi}{\pi}\right) \rangle_{q(\pi)}}\right)^{-1} \label{pip}\end{align}
Where note that $\langle \bm \theta \odot \bm \theta \rangle_{q_{\bm \psi}(\bm \theta)} = \bm \mu \odot \bm \mu + \bm \sigma \odot \bm \sigma$ using standard properties of multivariate Gaussians. Similarly, since $q(\pi) = \mc Beta(\pi|\xi_a, \xi_b)$ (see below), we have using the properties of the Beta distribution that for digamma function $\tilde \psi_0(\cdot)$:
\begin{align} \left\langle \log \left(\frac{1-\pi}{\pi}\right) \right\rangle_{q(\pi)} = \tilde \psi_0(\xi_b) - \tilde \psi_0(\xi_a)\label{elogpi}\end{align}
For the CAVI update for $q(\pi)$ we have: 
\begin{align}q(\pi) & \propto exp \left\{ \left \langle \log p(\bm \gamma,\pi) \right \rangle_{q(\bm \gamma)} \right\} \nonumber \\
 & \propto exp \left \{(a-1+\sum_{j=1}^d\langle \gamma_j \rangle_{q(\gamma_j)})\log(\pi) +(b-1+d-\sum_{j=1}^d\langle \gamma_j \rangle_{q(\gamma_j)})\log(1-\pi) \right \} \nonumber \\
& \propto \mc Beta\left(\pi|a+\sum_{j=1}^d\langle \gamma_j \rangle_{q(\gamma_j)}, b+d-\sum_{j=1}^d\langle \gamma_j \rangle_{q(\gamma_j)}\right) \nonumber \\
& \propto \mc Beta\left(\pi|a+d \bar \lambda, b+d(1-\bar \lambda)\right) \quad : \quad \bar \lambda = \frac{1}{d} \sum_{j=1}^d \lambda \end{align}
Therefore $\bm \xi = (\xi_a, \xi_b) = \left(a+\sum_{j=1}^d\langle \gamma_j \rangle_{q(\gamma_j)}, b+d-\sum_{j=1}^d\langle \gamma_j \rangle_{q(\gamma_j)}\right)$ and the expected sufficient statistic $\langle \log \frac{1-\pi}{\pi} \rangle_{q(\pi)}$ is given by expression \ref{elogpi} above. 

\subsection{SSVGP Leave-one-out Predictive Density (LOOPD)}

Here we briefly derive the mathematical form of the leave-one-out predictive density under a given model $\mc M_k$ in the BMA ensemble under the zero-temperature posterior. The easiest way to do this is to note that model $\mc M_k$ is defined by parameters $\bm  \nu_k = (\bm \mu_k, \bm \phi_k, a,b,c, v_k)$. Note here we treat inverse lengthscales $\bm \theta$ as fixed at posterior means $\bm \mu_k$, since this is equivalent to using a zero-temperature posterior $q_{\bm \psi}(\bm \theta) = \prod_{j=1}^d\delta_{\mu_j}(\theta_j)$. Thus, since kernel parameters are fixed the LOOPD for model $k$ is simply that of a standard GP evaluated at $(\bm \theta = \bm \mu_k, \bm \phi= \bm \phi_k)$ \cite{rasmussen2003gaussian} (note we assume a zero mean GP prior as typical):
\begin{align}\prod_{i=1}^n p(y_i|\mc D_{\neg i}, \bm x_i, \mc M_k) & =  \prod_{i=1}^n p_{\bm \phi_k,\theta_k}(y_i|\mc D_{\neg i}, \bm x_i)\\
 & = \prod_{i=1}^n \mc N(y_i|\mu_i,\sigma^2_i) \\
 \mu_i & = \bm k_{X_{\neg i}}(\bm x_i)^T\left(K_{X_{\neg i}X_{\neg i}}+\sigma^2I\right)^{-1}\bm y_{\neg i} \\
 \sigma^2_i & = k_{\bm \mu_k, \tau_k}(\bm x_i,\bm x_i) - \bm k_{X_{\neg i}}(\bm x_i)^T\left(K_{X_{\neg i}X_{\neg i}}+\sigma^2I\right)^{-1}\bm k_{X_{\neg i}}(\bm x_i) + \sigma^2 \end{align}
 Here $K_{X_{\neg i}X_{\neg i}}$ is the $(n-1) \times (n-1)$ gram matrix of training points when removing ($y_i, \bm x_i)$, and $\bm k_{X_{\neg i}}(\bm x_i)$ is an $n-1$ dimensional vector of kernel function evaluations between $\bm x_i$ and all other training points. As discussed in the main text we use the algorithm in \citet{burkner2021efficient} in small-$n$ scenarios ($\mc O(n^3)$ complexity) and the nearest neighbour truncation algorithm in \citet{jankowiak2021scalable} in large-$n$ scenarios ($\mc O(nf(n)d)$ complexity with $f(n) \in [log(n),n]$). In some circumstances we add a small regularisation term  of $0  < \kappa \ll1$ to the posterior variances small-$n$ scenarios to prevent outliers from blowing up the negative log LOOPD. We only find it necessary to do this in the toy example (setting $\kappa = 0.1$), but in all main experiments performance is robust to $\kappa \in [0,1]$.

\subsection{SSVGP Predictive Distribution}

Here we derive the closed form predictive moments given by the predictive distribution approximations in eq. \ref{pred} and \ref{pred0} in the main text. Since both distributions are discrete mixtures of standard GP posterior predictives with the $k^{th}$ component parameterised by inverse lengthscales $\bm \theta_{k}$ kernel scale and noise parameters $\bm \phi_{k}$, we derive the moments for this distributional family. However note that for eq. \ref{pred} in the main text (i.e. the SSVGP posterior predictive with Mean-Field Gaussian $q_{\bm \psi}(\bm \theta)$ and fixed $v$), the distribution is an approximation to the true predictive posterior, and thus the predictive covariance derived below is  a biased but consistent estimator (it contains a product of approximate expectations).

We define the distributional family we work with as follows:
\begin{align}p(\bm y_*|X_*, \mc D) =  \sum_{k=1}^Kp_{\bm \theta_k,\bm \phi_k}(\bm y_*|X_*,  \mc D)w_k\end{align}
Here $\{w_k\}_{k=1}^k$ are the mixture weights and $p_{\bm \theta_k,\bm \phi_k}(\bm y_*|X_*,  \mc D)$ is a standard GP posterior predictive given $(\bm \theta_k, \bm \phi_k)$. In this case, the posterior predictive posterior mean is computed as:
\begin{align} \bm m(X_*)  = \mathbb E[\bm y_*|X_*, \mc D] & = \int \bm y_*  \sum_{k=1}^Kp_{\bm \theta_k,\bm \phi_k}(\bm y_*|X_*,  \mc D)w_kd \bm y_* \nonumber \\
& =\sum_{k=1}^Kw_k\int \bm y_*  p_{\bm \theta_k,\bm \phi_k}(\bm y_*|X_*,  \mc D)d \bm y_* \nonumber \\
& =\sum_{k=1}^Kw_k\bm \xi_{\bm \theta_k, \bm \phi_k}(X_*) \label{postmean} \end{align}
Where $\bm \xi_{\bm \theta_k, \bm \phi_k}(X_*) = \mathbb E_{p_{\bm \theta_k,\bm \phi_k}(\bm y_*|X_*,  \mc D)}[\bm y_*]$ is the mean of standard GP posterior predictive component $p_{\bm \theta_k,\bm \phi_k}(\bm y_*|X_*, \bm \theta_k, \mc D)$  (see \citet{rasmussen2003gaussian} for details).

Similarly, using the definition of covariance: $\mathbb V[\bm x] =  \mathbb E[\bm x \bm x^T] - \mathbb E[\bm x]\mathbb E[\bm x]^T$, the posterior covariance is given as:
\begin{align}\bm  V(X_*) =  \mathbb V[\bm y_*|X_*, \mc D] & = \sum_{k=1}^Kw_k\int \bm y_*\bm y_*^T  p_{\bm \theta_k,\bm \phi_k}(\bm y_*|X_*,  \mc D)d \bm y_* - \bm m(X_*)\bm m(X_*)^T \nonumber \\
& =\sum_{k=1}^Kw_k\left(\bm \Sigma_{\bm \theta_k, \bm \phi_k}(X_*)+ \xi_{\bm \theta_k, \bm \phi_k}(X_*)\xi_{\bm \theta_k, \bm \phi_k}(X_*)^T\right) - \bm m(X_*)\bm m(X_*)^T \end{align}
Where $\bm \Sigma_{\bm \theta_k, \bm \phi_k}(X_*)= \mathbb Var_{p_{\bm \theta_k,\bm \phi_k}(\bm y_*|X_*,  \mc D)}[\bm y_*]$ is the covariance of standard GP posterior predictive component $p_{\bm \theta_k,\bm \phi_k}(\bm y_*|X_*,\bm \theta_k, \mc D)$ (see \citet{rasmussen2003gaussian} for details)

\subsection{SSVGP Posterior Point of Intersection (PPI)}

Here we derive equation \ref{PPI} in the main text for the posterior point of intersection (PPI). The PPI is given by $\hat \theta = \sqrt{\langle \theta^2 \rangle_{q_{\bm \psi_j}(\theta)}}$ such that $q(\gamma = 1) = q(\gamma = 0) = \frac{1}{2}$. To derive the expression we substitute $\lambda = \frac{1}{2}$ into the optimal CAVI update for $\lambda$ given by equation \ref{pip}: 
\begin{align}\frac{1}{2} =  \left(1+\left(\frac{1}{c}\right)^{\frac{1}{2}}e^{-\frac{1}{2}\langle \theta_j^2 \rangle_{q_{\bm \psi_j}(\theta_j)}v(1-c)+ \langle \log \left(\frac{1-\pi}{\pi}\right) \rangle_{q(\pi)}}\right)^{-1}\end{align}
Re-arranging terms we get: 
\begin{align} 1 & = \left(\frac{1}{c}\right)^{\frac{1}{2}}e^{-\frac{1}{2}\langle \theta_j^2 \rangle_{q_{\bm \psi_j}(\theta_j)}v(1-c)+ \langle \log \left(\frac{1-\pi}{\pi}\right) \rangle_{q(\pi)}} \\
0 & = \frac{1}{2}\log \left(\frac{1}{c}\right)-\frac{1}{2}\langle \theta_j^2 \rangle_{q_{\bm \psi_j}(\theta_j)}v(1-c)+ \left\langle \log \left(\frac{1-\pi}{\pi}\right) \right\rangle_{q(\pi)} \end{align}
Taking the expected sufficient statistic of $\theta_j$ to the LHS and simplifying we get:
\begin{align}\langle \theta_j^2 \rangle_{q_{\bm \psi_j}(\theta_j)}v(1-c)= \log \left(\frac{1}{c}\right)+ 2\left\langle \log \left(\frac{1-\pi}{\pi}\right) \right\rangle_{q(\pi)} \end{align}
Thus finally we arrive at the equation in the main text:
\begin{align}\sqrt{\langle \theta_j^2 \rangle_{q_{\bm \psi_j}(\theta_j)}} = \hat \theta = \sqrt{\frac{\log\left(\frac{1}{c}\right)+ 2\langle \log \frac{1-\pi}{\pi} \rangle_{q(\pi)}}{v(1-c)}}\end{align}

\subsection{Free Energy Fixed Variable Inclusion Cost}

In Section 4 we discuss that the evidence lower bound has a fixed variable  `inclusion cost' of order $\log (\frac{1}{c})$, making it unideal for model selection as $c$ can heavily impact the degree of preference for sparse models. Here we derive the source of this inclusion cost. Starting from the KL regularisation term  of the evidence lower bound, note that under and arbitrary inverse lengthscale posterior $q_{\bm \psi}(\bm \theta)$, if we retain only the terms that include $\{\langle  \theta_j^2 \rangle_{q_{\bm \psi_j}(\theta_j)}\}_{j=1}^d$ and $\{\lambda_j\}_{j=1}^d$, we are left with: 
\begin{align}-KL[q_{\bm \psi}(\bm \theta)q(\bm \gamma)q(\pi)||p(\bm \theta, \bm \gamma, \pi)] & = \frac{1}{2}\sum_{j=1}^d \lambda_j \log(c)-\frac{v}{2}\sum_{j=1}^d \langle \theta_j^2 \rangle_{q_{\bm \psi_j}(\theta_j)}(\lambda_j c+1-\lambda_j) \nonumber \\
    & \quad \quad + \sum_{j=1}^d \lambda_j\left \langle \log\left(\frac{\pi}{\lambda_j}\right)\right \rangle_{q(\pi)}+\sum^d_{j=1}(1-\lambda_j)\left \langle \log\left(\frac{1-\pi}{1-\lambda_j}\right)\right \rangle_{q(\pi)} \nonumber\\
    & \quad \quad  + const. wrt.(\{\langle  \theta_j^2 \rangle_{q_{\bm \psi_j}(\theta_j)},\lambda_j\}_{j=1}^d) \end{align}
The contribution of variable $j$ to these components of the KL is given by: 
    \begin{align}\zeta_j = \lambda_j\frac{1}{2}\log(c) -\frac{v}{2}\langle \theta_j^2 \rangle_{q_{\bm \psi_j}(\theta_j)}(\lambda_j c+1-\lambda_j)+  \lambda_j\left \langle \log\left(\frac{\pi}{\lambda_j}\right)\right \rangle_{q(\pi)}+(1-\lambda_j)\left \langle \log\left(\frac{1-\pi}{1-\lambda_j}\right)\right \rangle_{q(\pi)}\end{align}
Clearly when moving from $\lambda_j:0 \to 1$, there is a fixed inclusion cost from the first term of $\frac{1}{2}\log\left( \frac{1}{c}\right)$. As a result the value of $c$ can have a substantial effect on whether $\mc F$ is higher when the average PIP is higher, which we find experimentally. This makes it challenging to use $\mc F$ reliably to do discrete model selection or averaging.

\subsection{Dirac Spike with Paired Mean Field (PMF) Approximation}\label{PMF}
In Section 4 we also implement a spike and slab GP using a Dirac point mass at zero for the spike prior distribution, and the paired mean field variational approximation used in \citet{dai2015spike} and \citet{carbonetto2012scalable}:
\begin{align}q(\bm \theta|\bm \gamma)q(\bm \gamma) = \prod_{d=1}^d [\mc N(\theta_j|\mu_j,\sigma^2_j)\lambda_j]^{\gamma_j}[\delta_{0}(\theta_j)(1-\lambda_j)]^{1-\gamma_j}\end{align}
This factorisation has the attractive property of retaining posterior dependence between the $j^{th}$ inclusion variable $\gamma_j$ and inverse lengthscale $\theta_j$. However the cost of this dependency is that exact CAVI updates for $q(\bm \gamma)$ are no longer tractable as it requires computing expectations of the likelihood term $\log p(\bm y|\bm \theta)$, which are intractable for most kernel functions: 
\begin{align}q( \gamma_j) & \propto \exp \left( \langle \log p(\bm y|\bm \theta) \rangle_{q(\bm \theta|\bm \gamma)q(\bm \gamma_{\neg j})}+ \langle \log p (\theta_j, \gamma_j) \rangle_{q(\theta_j| \gamma_j)} \right)\end{align}
Note that even if the expectations $ \langle \log p(\bm y|\bm \theta) \rangle_{q(\bm \theta|\bm \gamma)}$ could be computed for a setting of $\bm \gamma$, we would need to compute $2^{d-1}$ of these quantities. The complexity of this CAVI update would therefore be $\mc O(2^dn^3)$, which is computationally infeasible for large-$n$ or even moderate $d$.

Our implementation therefore optimises $q(\bm \theta, \bm \gamma)$ using stochastic gradient variational inference. We use score-gradients (aka BBVI \cite{ranganath2014black}) to estimate $\nabla_{\bm \lambda} \mc F$ (since $\bm \gamma$ are non-differentiable) and reparameterisation gradients (aka repgrad \cite{kingma2013auto}) to estimate $\nabla_{\bm \psi}\mc F$:
\begin{align} \nabla_{\bm \psi}\mc F \approx \frac{1}{S_0}\sum_{s=1}^{S_0} \nabla_{\bm \theta}\log p(\bm y|\bm \theta^{(s)})\Big|_{\bm \theta^{(s)} = \mc T_{\bm \psi}(\bm \epsilon^{(s)}, \bm z^{(s)}) }\nabla_{\bm \psi} \mc T_{\bm \psi}(\bm \epsilon^{(s)}, \bm z^{(s)}) - \nabla_{\bm \psi} KL[ q(\bm \theta|\bm \gamma)q(\bm \gamma)||p(\bm \theta, \bm \gamma)] \\
\nabla_{\bm \lambda }\mc F \approx \frac{1}{S_1}\sum_{s=1}^{S_1} \left(\log p(\bm y|\bm \theta^{(s)}) \nabla_{\bm \lambda} \log q(\bm \theta^{(s)})\right)\Big|_{\bm \theta^{(s)} = \mc T_{\bm \psi}(\bm \epsilon^{(s)}, \bm z^{(s)}) } - \nabla_{\bm \lambda} KL[ q(\bm \theta|\bm \gamma)q(\bm \gamma)||p(\bm \theta, \bm \gamma)] \end{align}
Note that now the differentiable transformation is $\mc T_{\bm \psi}(\bm \epsilon, \bm z) = \delta_{\bm 1}(\bm z \leq \bm \lambda) \odot \left(\bm \epsilon \odot \bm \sigma + \bm \mu\right)$, whereby $\bm z \sim \mc U[0,1]^d$ and $\bm \epsilon \sim \mc N(\bm 0, I)$ are drawn from the base distributions. Also note that the marginal distribution over inverse lengthscales is $q_{\bm \psi}(\bm \theta) = \prod_{j=1}^d\left[\lambda_j \mc N(\theta_j|\mu_j, \sigma^2_j) +(1-\lambda_j)\delta_0(\theta_j)\right]$. In practice we reparameterise the model in terms of $\bm \kappa \in \mathbb R^d$ where $\bm \lambda = (\bm 1 - \exp(-\bm \kappa))^{-1}$ and optimise $\bm \kappa$. We set $S_0=1$ consistent with our a-CAVI algorithm, but set $S_1 \in \{10,100\}$ (i.e. two implementations used) due to the much higher degree of score-gradient variance. We treat $(\pi, v)$ as fixed parameters but pre-tune these to optimise performance for comparison against our SSVGP with the a-CAVI algorithm. See Appendix B.1 for implementation details.

\subsection{Proposition and Proof of a-CAVI Shrinkage Property}

In this section we present a shrinkage property of a-CAVI Algorithm 1 along with a simple proof. We subsequently discuss how this property can be used to motivate the dropout pruning procedure we introduce in Section 4 of the main text. For simplicity our proof is for a 1d-input scenario but can be extended to multiple input dimensions.

The conditions required for this property to hold only apply under a simplification of the a-CAVI algorithm. However they mathematically formalise the notion that for a particular range of the spike and slab parameter $v$, small enough PIP values $\lambda_j <\epsilon \in (0,1)$ can enforce that $\mu_j$ converges to an $\delta$-neighbourhood of zero.

For the proof we require boundedness and Lipschitz continuity on the gradient of the log likelihood $\nabla_\theta \log p(\bm y| \theta)$. We define these assumptions as follows: 

\begin{definition}

For a real valued differentiable function $f:\mc X \to \mathbb R$ where $\mc X \subseteq \mathbb R^d, d \in \mathbb N_{>0}$, the gradient $\nabla f(\bm x)$ is $B$-bounded if $\exists B\in \mathbb R_{\geq0}$ such that $||\nabla f(\bm x)||_2 \leq B \quad \forall \bm x \in \mc X$.
\end{definition}

\begin{definition}

For a real valued differentiable function $f:\mc X \to \mathbb R$ where $\mc X \subseteq \mathbb R^d, d \in \mathbb N_{>0}$, the gradient $\nabla f(x)$ is $C$-Lipschitz continuous if $ \exists C\in \mathbb R_{\geq 0}$ such that $||\nabla f(\bm x)-\nabla f(\bm y)||_2 \leq C||\bm x - \bm y||_2 \quad \forall \bm x,\bm y \in \mc X$. 

\end{definition}
Going forward we use $C$-smoothness to denote $C$-Lipschitz continuity for convenience.
\begin{proposition}
\label{prop1}

Consider the spike and slab Gaussian process model defined in Section 3 of the main text where $\bm y \in \mathbb R^{n}, X \in \mathbb R^{n \times 1}$ is the dataset, $\theta \in \mathbb R$ is the inverse lengthscale (where parameterisation is chosen to enable $k_\theta(x,x') = k_{-\theta}(x,x')$ - see Appendix \ref{kernel_param} for details), $\bm \phi = (\tau, \sigma^2) \in \mathbb R^2_{>0}$ are kernel scale and noise parameters, $\gamma \in \{0,1\}$ is the latent binary inclusion variable, $\pi \in [0,1]$, and spike and slab prior is parameterised such that $(v,c) \in \mathbb R^2_{>0}$ with $0 < c \ll1$. Let the a-CAVI algorithm be defined by the following simplifications. We use zero-temperature posterior $q_{\bm \psi_j}(\theta) = \delta_{\mu}(\theta)$ such that $\theta = \mu$, let the prior inclusion probability $\pi$ be fixed at $\pi_0 \in(0,1)$\footnote{This can be simulated by setting Beta prior hyperparameters to $a  = \tau$, $b =\tau\left(\frac{1}{\pi_0}-1\right)$ and letting $\tau \to \infty$.}, fix $\bm \phi = \hat {\bm \phi} \in \mathbb R^2_{>0}$ and use a constant learning rate $\eta \in \mathbb R_+$. Let the log likelihood gradient $\nabla_\theta\log p(\bm y|\theta)$ be $B$-bounded and $C$-smooth. 

Then, when running a-CAVI for $T>1$ iterations, if at any iteration $t \in \{1,...,T-1\}$ we get $\lambda^{(t)}\leq \epsilon$, and $\exists \delta>0, \kappa > 0$, $\epsilon \in (0,1)$ such that: 

\begin{align}
 \frac{B+\kappa}{\delta(1-\epsilon)}\leq v & \leq \frac{2}{\delta^2(1-c)}\log\left(\frac{\epsilon(1-\pi_0)}{c^{\frac{1}{2}}(1-\epsilon)\pi_0} \right) \label{p0a} \\
0 < \eta & \leq \frac{\alpha}{B+v(\delta+\alpha)} \quad \forall \alpha \in [\min(\delta,\frac{\kappa}{C+v}),\infty) \label{p0b}
\end{align}
we have that as $T \to \infty$:  $|\mu^{(T)}| \to |\mu^\infty| \leq \delta$.
\end{proposition}

\begin{proof}
\label{proof1}
We prove this by showing that under the assumptions above$\text{ } \forall j\geq 1$: $|\mu^{(t+j)}|< |\mu^{(t+j-1)}|$ if $|\mu^{(t+j-1)}|>\delta$, and $|\mu^{(t+j)}|\leq\delta$ otherwise, which implies $\lambda^{(t+j)} \leq \epsilon$. Taking $j \to \infty$ we recover the result using bounds on $\nabla _\mu \mc F$ when $|\mu|>\delta$ that ensures the sequence $\{\mu^{(t+j)}\}_{j=1}^\infty$ converges to $\mu^\infty \in [-\delta,\delta]$. Note $\mu^{(t+1)}$ is obtained after running $S$ gradient-steps at a-CAVI iteration $t+1$, and $\lambda(\cdot)$ is the optimal CAVI update function in equation \ref{pip}.    \\

We first show that if $\lambda^{(t)} \leq \epsilon$ and $|\mu^{(t)}|> \delta$ we must have that $\text{sign}(\nabla_\mu \mc F(\mu^{(t)},\lambda^{(t)}))= -\text{sign}(\mu^{(t)})$. Starting with the LHS inequality of condition \ref{p0a}, using $B$-boundedness of the log-likelihood gradient and that due to our kernel parameterisation $\nabla_\mu \mc F(\mu,\cdot) = -\nabla_{\mu}\mc F(-\mu, \cdot)$, we have
\begin{align}
    v(1-\epsilon) & \geq \sup_{\theta \in (\delta, \infty)} \left\{\frac{\nabla_\theta \log p(\bm y|\theta)}{\theta}\right\}+ \frac{\kappa}{\delta}\\
    & =  \sup_{|\theta| \in (\delta, \infty)} \left\{\frac{\nabla_\theta \log p(\bm y|\theta)}{\theta}\right\}+\frac{\kappa}{\delta}\\
    & > \frac{\nabla_\theta \log p(\bm y|\theta) + (-1)^{\mathbb I(\theta<0)}\kappa}{\theta}\quad \forall |\theta| \in (\delta, \infty)\label{p0}
\end{align}
where $\mathbb I(\cdot)$ is the indicator function. Substituting in $v(1-\lambda^{(t)}+\lambda^{(t)}c) > v(1-\epsilon)$ on the LHS, multiplying by $\theta$ and re-arranging, we get $\nabla_\mu \mc F(\mu^{(t)},\lambda^{(t)}) < -\kappa$ when $\mu^{(t)}>\delta$ and  $\nabla_\mu \mc F(\mu^{(t)},\lambda^{(t)}) > \kappa$ when $\mu^{(t)}<-\delta$. This means the gradient always points to the origin when $|\mu^{(t)}|> \delta$.  \\

We now show that for a small enough learning rate $\eta$ we can guarantee a single gradient step in $\mu$ will strictly decrease $|\mu^{(t)}|$ if $|\mu^{(t)}|>\delta$. To achieve this we require that: 
\begin{align}
|\mu^{(t+1})| < |\mu^{(t)}+\eta \nabla_\mu \mc F(\mu^{(t)}, \lambda^{(t)})|
\end{align}
Starting with the positive case $\mu^{(t)} = \delta + \alpha : \alpha > 0$, due to the strictly negative gradient we only need
\begin{align}
-(\delta + \alpha) < \delta + \alpha +\eta \nabla_\mu \mc F(\delta+ \alpha, \lambda^{(t)})
\end{align}
Re-arranging and using the strictly negative gradient we get: 
\begin{align}
\eta < \frac{2(\delta + \alpha)}{|\nabla_\mu \mc F(\delta+ \alpha, \lambda^{(t)})|}
\end{align}
Now, since the log-likelihood gradient is $B$-bounded note that $\nabla_{\mu\in [-a,a]} \mc F$ is $(B+va)$-bounded, where $a>0$. Thus we need
\begin{align}
\eta < \frac{2(\delta + \alpha)}{B+v(\delta+\alpha)} \forall \alpha \geq 0
\end{align}
Which is satisfied by assumption \ref{p0b}.\\

The same result holds by symmetry when examining the negative case $\mu^{(t)}=-(\delta + \alpha)$. By implication, every gradient step $s = 1,...,S$ taken during a-CAVI iteration $t+1$ to update $\mu^{(t)} = \mu^{(t)}_1 \to \mu^{(t)}_S =  \mu^{(t+1)}$, must strictly decrease $|\mu^{(t)}_s|$ at any step $s$ where $|\mu^{(t)}_s| \geq \delta$. \\

Now we show that if $|\mu^{(t)}|\leq \delta$ then so is $|\mu^{(t+1)}|$. For the positive case, denote $\mu^{(t)} = \delta - \alpha : \alpha \in [0,\delta]$ For the result, we need to guarantee:
\begin{align}
     \delta - \alpha + \eta \nabla_\mu \mc F(\delta - \alpha ,\lambda^{(t)})  \in [-\delta, \delta] \label{p1}
\end{align}
Now note that since the log-likelihood gradient is $C$-smooth, then $\nabla_\mu \mc F$ must satisfy $(C+v)$-smoothness (from the gradient of the KL term). Thus, we have that $\nabla_\mu \mc F(\delta - \alpha ,\lambda^{(t)}) \in [-B-v(\delta-\alpha),\min(B+v(\delta-\alpha),-\kappa + \alpha (C+v)] \text{ } \forall \alpha \in [0,\delta]$. The $\min(\cdot, \cdot)$ accounts for the boundedness of $\mc \nabla \mc F$ when $\mu \in [-\delta, \delta]$. \\

Using this condition, for the lower bound of eq. \ref{p1} to hold we need:
\begin{align}
     \delta - \alpha - \eta(B+v(\delta-\alpha))  \geq -\delta
\end{align}
Which implies
\begin{align}
     \eta \leq \frac{2\delta - \alpha}{B+v(\delta - \alpha)} \forall \alpha \in [0,\delta] \label{p2}
\end{align}
For the upper bound to hold we need to guarantee 
\begin{align}
     \delta - \alpha + \eta\min(B+v(\delta - \alpha),-\kappa+\alpha (C+v))  \leq \delta
\end{align}
Which implies (noting we only need to consider the case where the gradient is strictly positive):
\begin{align}
     \eta \leq \frac{\alpha}{\min(B+v(\delta - \alpha),-\kappa+\alpha (C+v))} \forall \alpha \in (\min(\frac{\kappa}{C},\delta),\delta] \label{p3}
\end{align}

Both eqs \ref{p2} and \ref{p3} are clearly satisfied by assumption \ref{p0b}. By symmetry of the problem, when we consider the negative case $\mu \in [-\delta, 0]$ we recover the same inequalities. \\

The above results mean that we must have
\begin{align}
    |\mu^{(t+1)}|< \max(|\mu^{(t)}|, \delta+\beta) \quad \forall \beta>0 \label{p4}
\end{align} 

Now, since $\lambda^{(t)} \leq \epsilon$, note that if \ref{p4} holds we must also have 
\begin{align}
    \lambda^{(t+1)}< \max(\lambda^{(t)},\lambda(\delta)+\beta) \quad \forall \beta > 0 \label{p5}
\end{align}
Where $\lambda(\cdot)$ is the optimal CAVI update used given by equation \ref{pip}. The last line simply uses that under the assumptions in this proposition $\lambda'(\cdot)> 0$ everywhere. \\

From the RHS inequality of condition \ref{p0a} and using \ref{p5} we also have $\lambda^{(t+1)}\leq \epsilon$, since rearranging this inequality in \ref{p0a} we get $\lambda(\delta) \leq \epsilon$. Thus, the same results hold for iteration $t+j$ $\forall j \geq 1$ and so we can write:
\begin{align}
    |\mu^{(t+j)}| & < \max(|\mu^{(t+j-1)}|, \delta+\beta) \quad  \forall \beta>0, \forall j\geq 1 \\
    \lambda^{(t+j)} & \leq \epsilon  \quad  \forall j\geq 1 \
\end{align}
Since when $|\mu| > \delta$ we have by eq. \ref{p0} that $|\nabla_\mu \mc F|> \kappa$ and $\nabla_\mu \mc F$ points to the origin, we can write that:
\begin{align}
    |\mu^{(t+j)}| & \leq |\mu^{(t)}| -j\eta\kappa \quad  \forall j\geq 1, \forall |\mu^{(t)}|>\delta
\end{align}
By taking $j,T \to \infty$ we recover $|\mu^{\infty}|\leq \delta$. In other words, under the assumptions in the proposition, the gradient outside of the $\delta$-neighbourhood of zero is always large enough to get from $\mu^{(t)} \in \mathbb R$ to the $\delta$-neighbourhood in infinite computation time, if $\lambda^{(t)} \leq \epsilon$.
\end{proof}

\begin{remark}
 Proposition \ref{prop1} states that for a certain range of spike precision values and small enough learning rate $\eta$, if $\lambda$ is set small enough at any point during a-CAVI iterations, the algorithm is guaranteed to converge the inverse lengthscale parameter $\mu$ to an $\delta$-neighbourhood of zero in infinite computation time. In particular, the spike precision must be large enough to dominate the gradient information in the log-likelihood for suitably small $\lambda$, but small enough to guarantee $\lambda(\mu) \leq \epsilon$ for any $|\mu|\leq \delta$. Notably, no assumptions have been made on the data generating process here outside of the boundedness and smoothness of the log-likelihood gradient. 
\end{remark}

\begin{remark}
 Proposition \ref{prop1} supports our use of dropout pruning, as it suggests that under certain conditions, permanently enforcing $\mu = 0$ only deviates our found local optima within a $\delta$-neighbourhood, thus enabling us to enjoy its computational speed ups without notable performance loss. 
 \end{remark}

\begin{remark}
We note that the proposition becomes less likely to hold as $n$ grows when using nearest neighbour minibatching, as the expected likelihood term contains a $\frac{n}{m}$ rescaling factor. In fact, this rescaling factor means that as $n\to \infty$ the recovered parameters ($\bm \mu, \bm \phi$) under ZT posterior $q_{\bm \psi}(\bm \theta)$ should be no different than those recovered using the minibatch SGD procedure of \citet{chen2020stochastic} i.e. the learned PIPs have no effect on $q_{\bm \psi}(\bm \theta)$. In this case we motivate dropout pruning from the perspective of reinstating the `message passing' link between $q(\bm \gamma)$ and $q_{\bm \psi}(\bm \theta)$, by augmenting the evidence lower bound with an L0 regularisation penalty on $\mu_j$ with penalty scale $\lim_{\tau \to \infty}\left(\tau\delta_{1}(\lambda_j \leq \epsilon)\right)$. 
\end{remark}

\section{EXPERIMENTAL AND IMPLEMENTATION DETAILS}

Here we discuss in detail the experiments ran and settings used for all methods implemented. All results were run on a 2.70GHz Intel(R) Core(TM) i7-10850H CPU with 32MB RAM. All code for our SSVGP is written in Python and is available at the Github repository {\color{blue}https://github.com/HWDance/SSVGP}. Note that barring the toy example, when we refer to the SSVGP, we refer to using our Bayesian model averaging (BMA) procedure introduced in Section 4 of the main paper, where models are trained using a-CAVI Algorithm 1 with the zero-temperature posterior $q_{\bm \psi}(\bm \theta)$ and dropout pruning procedure. The first toy example is used to demonstrate that our initial method (without BMA, zero-temperature and dropout) can perform well, but the adjustments are required for simultaneously reliable, fast and scalable performance. 

\subsection{Toy Example}

Here we consider the problem of identifying relevant variables in a high-dimensional synthetic dataset. We draw a univariate response from a sinusoidal function of 5 inputs corrupted with a small amount of noise: 
\begin{align} y = f(\bm x) + \epsilon \\
f(\bm x) = \sum\nolimits_{j=1}^5sin(a_jx_j) \\
\bm x \sim \mc N_d(0,I)\\
\epsilon \sim \mc N(0,\sigma^2)
\end{align}
We set $d=100$ with $\{a_j\}_{j=1}^5$ grid spaced over [0.5,1] and draw datasets of size $n=300$ using this generative process, controlling the noise variance to $\frac{1}{20}$ of the signal variance: $\sigma^2 = 0.05 Var[f(X)]$. Thus only 5/100 signals are relevant. Before running all methods we standardise $\mc D = (\bm y, X)$ to have mean zero and unit variance. 

Our initial implementation of the SSVGP on this example uses a Mean-Field Gaussian approximation for inverse lengthscale posterior: $q_{\bm \psi}(\bm \theta) = \mc N(\bm \mu, Diag(\sigma^2_1,...,\sigma^2_d))$ and fixed value of spike prior hyperparametrer $v$. We train using $K=5$ a-CAVI iterations with $T=200$ gradient steps for the first a-CAVI iteration, and $T=100$ steps thereafter. We use default ADAM smoothing parameters of $(\beta = 0.9, \beta_2=0.999)$ a learning rate of $\eta = 0.01$ and $S=1$ Monte Carlo samples in reparameterisation gradient approximations. Prior hyperparameters are set as $(a,b,c,v) = (10^{-3}, 10^{-3},10^{-8}, 10^4)$ so that the spike precision $v=10^4$ approximates a Dirac spike at zero, and the slab precision $cv = 10^{-4}$ approximates a uniform distribution. We initialise $\bm \lambda = \bm 1$ (this is critical to avoid shrinking inverse lengthscales for relevant variables in the initialisation phase due to the regularisation of the KL term), and set $\bm \xi = \bm 1$. We initialise inverse lengthscale posterior means to $\bm \mu = \bm 1 \times d^{-\frac{1}{2}}$ (this prevents numerical underflow in kernel evaluations in high-dimensional designs), and posterior variances to $\bm \sigma = 2 \times 10^{-3}$ and set $\bm \phi  = \bm 1$. We also add a jitter of $10^{-3}$ to the diagonals of $K_{XX}$.

The ML-II GP is implemented using the GPytorch \cite{gardner2018gpytorch} library for 1000 iterations of ADAM with a learning rate of 0.1 and the same initialisation as our SSVGP for shared parameters. We use our own implementation of the Dirac spike and slab GP with paired mean field (PMF) approximation since there is no existing implementation in GP libraries (see Appendix A for mathematical model details). We run two versions with $S = 10$ and $S=100$ Monte Carlo samples used respectively for score gradient estimation of $\nabla_{\bm \lambda} \mc F$. Since our implementation does not use control variates \cite{liu2019rao} or Rao-Blackwellisation \cite{salimans2014using} for variance reduction, we use the $S=100$ sample version as an indication of obtainable performance using $S=10$ samples were these techniques used, since \citet{mohamed2020monte} find these can techniques can reduce variance by close to an order of magnitude in some experiments. We also optimise performance by a-priori fixing the prior inclusion probability to the true value ($\pi = 0.05$) and pre-tuning the slab precision $v_1 \in \{10^{-1},10^{-2},10^{-3},10^{-4}, 10^{-5}\}$. We train using the same number of total gradient iterations (500) and initialisation (for shared parameters) as the SSVGP except we initialise $\bm \lambda = 0.5$, which performed better than $1$.

We subsequently re-implement the SSVGP (our final method) using the BMA procedure, zero-temperature posterior restriction and dropout pruning introduced in Section 4. We use 11 values of spike precision $v$ grid-spaced over $\in \mc V = 10^4 \times 2^{\{-log_2(1000),log_2(1000)\}}$ such that $v \in [10,10^7]$. We set the pruning threshold to $\epsilon = 0.5$ and use a slightly larger base learning rate of $\eta = 0.05$ (our default recommendation in small-$n$ settings for our final method). We find this performs better in the absence of gradient variance due to the zero-temperature assumption. We use $m=\frac{n}{4}$ minibatching to further reduce runtime, and as we find this does not harm performance in our experiments. No nearest neighbour truncation of the LOOPD and predictive distribution is used.

We complete 10 trials of each method on synthetic dataset draws, reporting results in Tables \ref{t1} and \ref{t3} in the main text (we include test performance on $n_* = 100$ test points drawn from the same process). Figures \ref{f1} and \ref{f2} in the main text display average size ordered profiles for parameters that can be used to infer variable relevance for different methods (e.g. inverse lengthscales for the ML-II GP and PIPs for the SSVGP implementations). These profiles are constructed by size ordering the recovered parameters within each trial and then averaging the resulting profiles over the 10 trials. However we constrain the first 5 variables in the ordering (LHS of black dashed line) to be selected from the 5 relevant variables. Thus, the $4^{th}$ column in the variable ranking corresponds to the average value of the $4^{th}$ largest parameter recovered amongst the relevant variables, and the $10^{th}$ column corresponds to the average value of the $5^{th}$ largest parameter recovered amongst the irrelevant variables. 

These profiles enable us to clearly visualise the ability of our methods to better distinguish between irrelevant and relevant variables correctly vs. the ML-II GP. In particular, we see that the ML-II GP produces a smooth decline in the profile with no real discontinuity at the true inclusion/exclusion threshold, and fails to shrink the inverse lengthscales to arbitrarily small values (e.g.$<0.01$) for the irrelevant dimensions (see Fig \ref{f1}), making it very challenging to use thresholding reliably to do variable selection. By contrast, our SSVGP shows a huge discontinuity both in PIPs and (posterior mean) inverse lengthscales at the true inclusion cut-off (see Fig. \ref{f2}), and also learns to include all relevant variables with PIP $\approx 1$ and include all irrelevant variables with PIP$\ll$0.5.

\subsubsection{Analysis of Posterior Point of Intersection (PPI)}

In Section 4 we demonstrated the sensitivity of the a-CAVI algorithm recovered PIPs to the value of spike and slab hyperparameter $v$. We also argued that the root of the problem is that the posterior point of intersection (PPI) is generally much more insensitive to $q(\pi)$ (which adapts the PPI during a-CAVI iterations) than to $v$. Recall that the PPI is the value of the transformed expected sufficient statistic of the inverse lengthscale $\hat \theta = \langle \theta^2 \rangle_{q_{\bm \psi_j}(\theta)}$ that induces an even probability of inclusion for variable $j$ in the CAVI update for $q(\gamma_j)$:
\begin{align}\hat \theta = \sqrt{\frac{\log\left(\frac{1}{c}\right)+ 2\langle \log \frac{1-\pi}{\pi} \rangle_{q(\pi)}}{v(1-c)}}\end{align}
 To demonstrate the relative sensitivity to $q(\pi)$ and $v$, below in Figure \ref{f_a0} we display how the PPI varies with $\pi_0$ and $v$ (under the restriction that $q(\pi) = \delta_{\pi_0}(\pi)$ for simplicity). As we can see, unless $\pi_0$ is extremely small or large, the PPI varies minimally over $\pi_0$ but varies substantially with $v$ (the PPI is $\mc O(v^{-\frac{1}{2}})$. Thus $v$ can entirely determine the sparsity of the learned PIPs $\bm \lambda$.

\begin{wrapfigure}[18]{r}{0.4\textwidth}
\centering
\includegraphics[width = 0.4\textwidth]{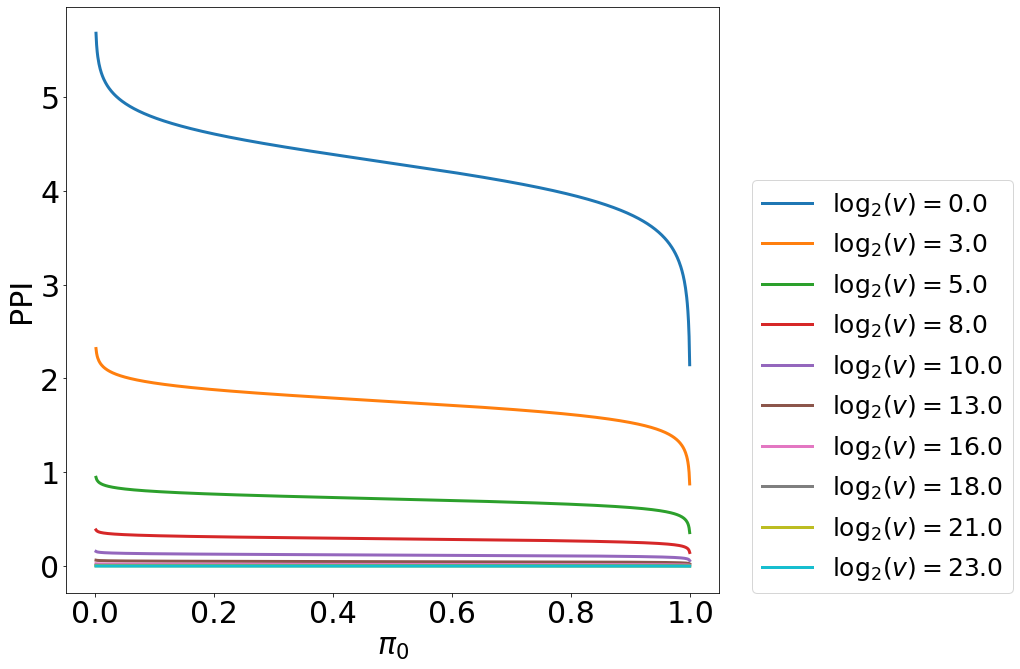}
\caption{Posterior point of intersection (PPI) over 10 values of $v$ grid spaced over $ 10^{\{0,...7\}}$ and 1000 values of $\pi_0$ grid spaced over $[ 0.001,0.999]$.}
\label{f_a0}
\end{wrapfigure}

 The other issue is that the range of PPI values that give good variable selection accuracy may vary based on the function properties, and so it is not the case that a single value of $v$ will perform well across all designs. For example, a function that varies much more slowly may require a lower PPI to prevent false exclusions of small but relevant inverse lengthscales. 
 
\begin{figure}[H]
     \centering
    \makebox[\textwidth][c]{\includegraphics[width = 0.3075\textwidth]{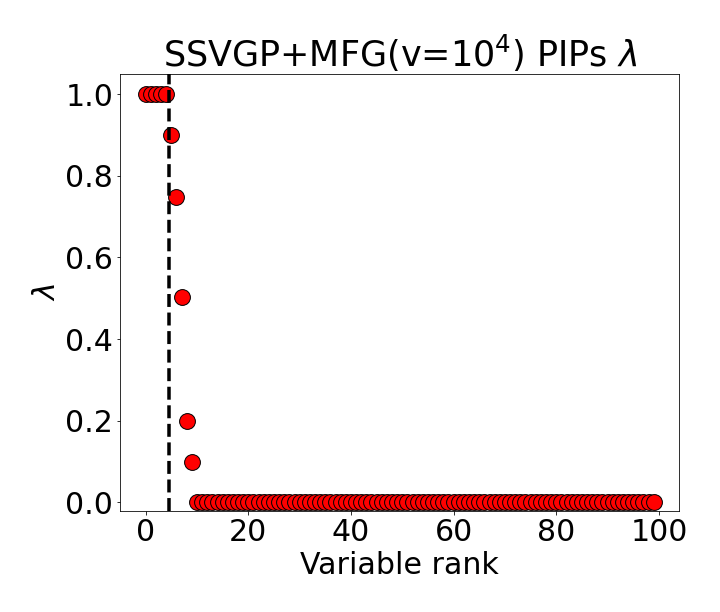}
        \includegraphics[width = 0.3075\textwidth]{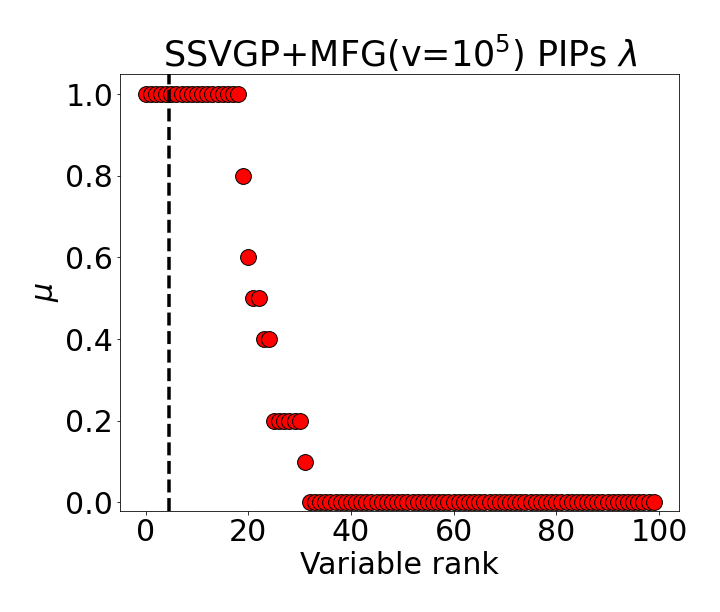}
    \includegraphics[width = 0.3\textwidth]{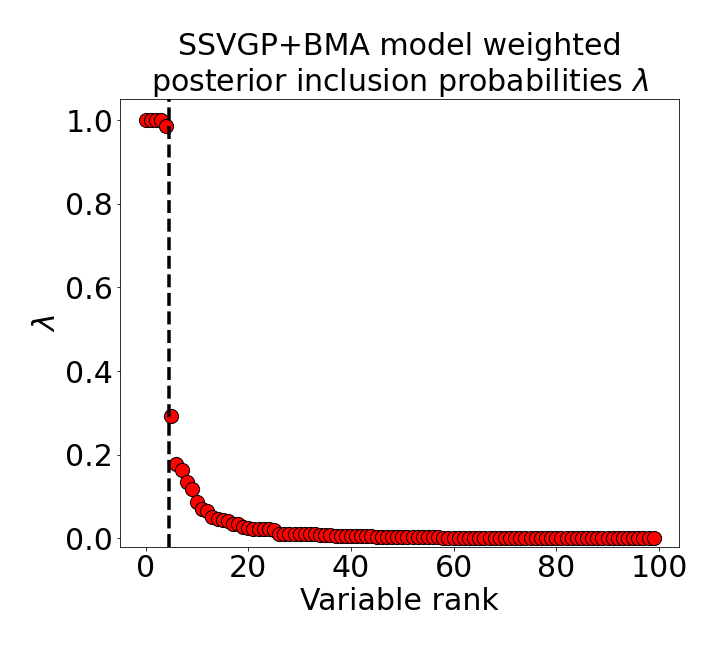}}
     \caption{Toy example size ordered PIP profiles (averaged over 10 trials) using the original design (5 relevant variables) for SSVGP with (LHS) Mean-Field Gaussian and fixed $v=10^4$, (Center) Mean-Field Gaussian and fixed $v=10^5$, (RHS) BMA with zero-temperature posterior (our final method). Profiles are constructed by size ordering the PIPs per trial with $q=5$ relevant variables constrained to take the first $q$ places in the ordering (LHS of black dashed line), and averaging the resulting profiles over the trials.  $v=10^4$ performs much better than $v=10^5$, but doing BMA over 11 values of $v$ grid spaced over $10^4 \times 2^{\{-log_2(1000),log_2(1000)\}}$ performs best. }
     \label{f_a1}
\end{figure}

\begin{figure}[H]
     \centering
    \makebox[\textwidth][c]{\includegraphics[width = 0.3075\textwidth]{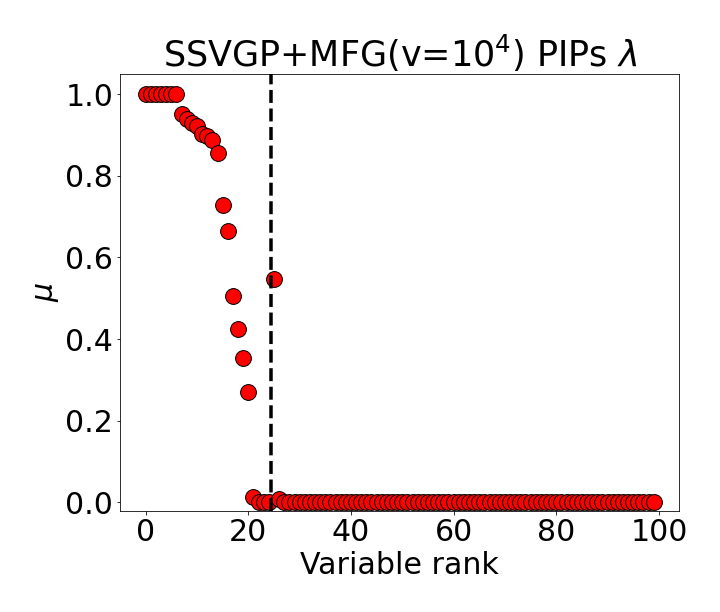}
    \includegraphics[width = 0.3075\textwidth]{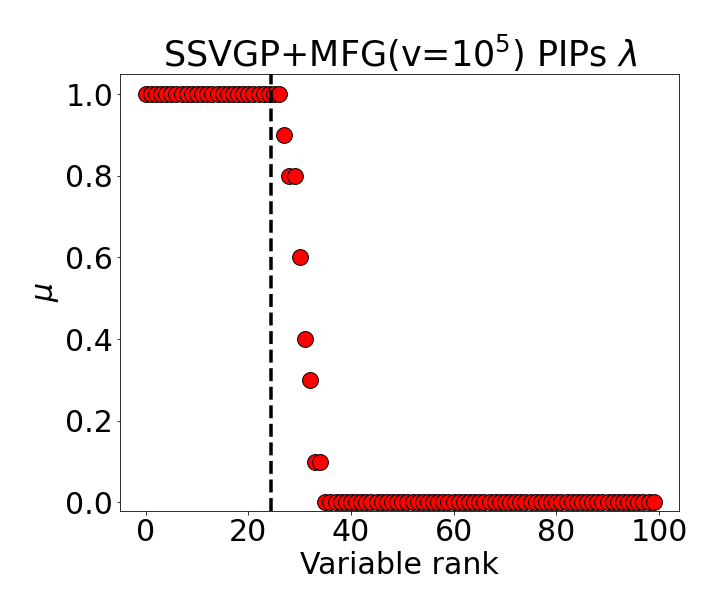}
    \includegraphics[width = 0.3\textwidth]{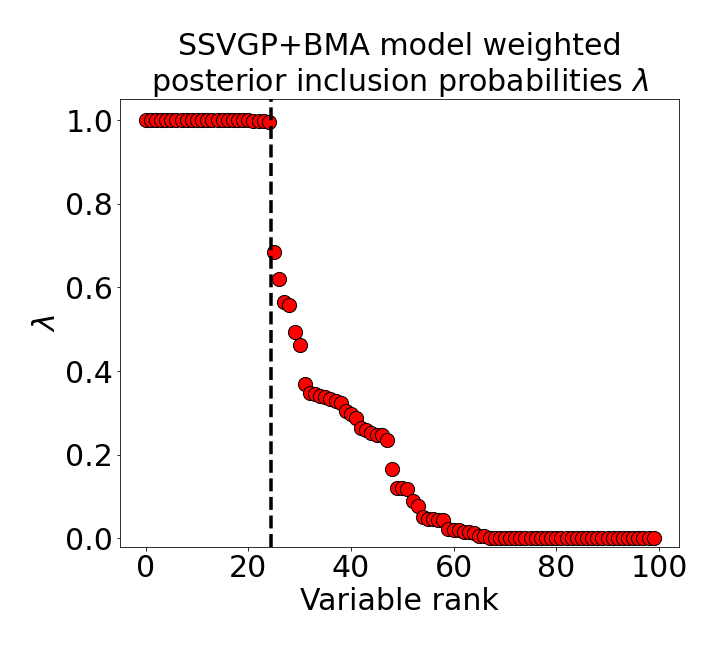}}
     \caption{Toy example average size ordered PIP profile using the modified design (25 relevant variables) for SSVGP with (LHS) Mean-Field Gaussian and fixed $v=10^4$, (Center) Mean-Field Gaussian and fixed $v=10^5$, (RHS) BMA with zero-temperature posterior. By comparison with Fig. \ref{f_a1}, $v=10^5$ performs better than $v=10^4$. However BMA performs best again, demonstrating its ability to effectively adapt to the generative process.}
     \label{f_a2}
 \end{figure}

 To demonstrate this, we display on Figure \ref{f_a1} below the PIP profiles recovered on the 10 trials of the toy example using our SSVGP with a Mean-Field Gaussian posterior, for two fixed values of $v$ ($10^4$ and $10^5$), as well as for our final method (SSVGP+BMA with zero-temperature and dropout pruning). We also display on Fig. \ref{f_a2} an equivalent set of profiles recovered on a modified version of the generative process where now 25/100 variables are relevant $\{a_j\}_{j=1}^{25}$ are grid spaced over [0.05,0.1]. Whilst $v=10^4$ performs well on the original design, but poorly on the second design, the opposite holds for $v=10^5$. However the BMA procedure is able to effectively adapt the PPI so that the PIPs more accurately reflect each generative design.
 
\subsection{Experiment 1}

We next test our SSVGP on 50 trials of the additive function used in Savitsky et al's (2011) main high-dimensional experiment. They use this experiment to test their MCMC algorithm for the spike and slab GP. Since spike and slab priors with a Dirac spike and MCMC for posterior inference is considered a gold-standard in Bayesian variable selection \cite{narisetty2014bayesian}, if our SSVGP can perform similarly well to their algorithm, we can be confident that our method can compete with best practice (but in a tiny fraction of the runtime). They were only able to complete a single trial of their method as they required $800,000$ MCMC iterations to ensure sufficient mixing of the chain (first $400,000$ discarded as burn-in). We note that they do not produce the runtime of their trial, but they do produce a runtime of $10224s$ when running the same algorithm for 500,000 iterations on an identically sized dataset ($n=100,d=1000$). Thus, we conclude with confidence that their runtime for the main high-dimensional experiment is likely to be $>10000s$.

The experiment draws $n=100$ training samples and $n_*=20$ test samples from the following generative process: 
\begin{align}
    y & = x_1+x_2+x_3+x_4+sin(3x_5)+sin(5x_6)+\epsilon \\
    \epsilon & \sim \mc N(0,0.05^2) \\
    \bm x & \sim \mc U[0,1]^{1000}
\end{align}
Thus, 6/1000 input variables are relevant. We implement our SSVGP (from now on with BMA, zero-temperature and pruning) and the ML-II GP using exactly the same implementation settings as in the toy example. We use three minibatch sizes of $m \in \{\frac{n}{4}, \frac{n}{2}, n\}$ to test how this affects performance. We also implement the following models with embedded variable selection: (LASSO \cite{tibshirani1996regression}, MCP \cite{zhang2010nearly}, GAM with group LASSO penalty \cite{huang2009group} and GAM with group spike and slab LASSO (SS LASSO) penalty \cite{bai2020spike}. The LASSO and MCP were implemented using the R \verb|ncvreg| package \cite{breheny2021package}, the GAM LASSO and GAM spike and slab LASSO (SSwere implemented using the R package \verb|sparseGAM| \cite{bai2021spike}. In all cases default cross-validation procedures were used except for the GAM SS LASSO where 3 fold CV was used to keep runtimes manageable. All code was ran using the \verb|rpy2| library in Python. We standardise inputs to have zero mean and unit variance following \citet{savitsky2011variable}.

Fig. \ref{f3} in the main text presents mean-squared error (MSE) and runtime performance percentiles for 50 trials of the experiment. We measure variable selection accuracy using the Matthews Correlation Coefficient (MCC $\in [-1,1]$) - a best practice measure of imbalanced classification accuracy \cite{chicco2020advantages}. We select using using a PIP threshold of $\bar \lambda = 0.5$ with the SSVGP. For the ML-II GP we report MCC performance using the \emph{best} performing threshold from $0.1^{0,0.5,1,1.5,2}$, as indicative 'best-case' performance achievable with inverse lengthscale thresholding. All other methods automatically do binary variable selection. We report performance percentiles instead of mean and standard deviation as we found the distribution to be heavily skewed by a few outlier trials with poor performance. Detailed performance for different percentiles is contained in Table \ref{t_a1} below. For transparency, we additionally display the mean and standard deviation performance below in Table \ref{t_a2}, but note the large standard deviations reflecting the outliers. We normalise the mean-squared error by the variance of the response as in Savitsky et al (2011).

\begin{table}[H]
    \centering
    \begin{tabular}{lrrr|rrr|rrr}
\hline
& \multicolumn{3}{c}{$25^{th}$ percentile} & \multicolumn{3}{c}{$50^{th}$ percentile} & \multicolumn{3}{c}{$75^{th}$ percentile}\\
                      &    MSE &   MCC & Runtime &    MSE &   MCC & Runtime &    MSE &   MCC & Runtime   \\
\hline
 SSVGP(m=n)           & 0.0046 & \textbf{1}    &      19.6  & 0.0082 &  \textbf{1}    & 19.9    & 0.0133 & 0.93 &      20.5  \\
 SSVGP(m=n/2)         & \textbf{0.0043} & \textbf{1}    &      10.4 & \textbf{0.0068} &  \textbf{1}    & 10.6   & \textbf{0.0111} & \textbf{1}    &      11.1     \\
 SSVGP(m=n/4)         & 0.005  & \textbf{1}    &       7.9 & 0.0086 &  \textbf{1}    & 8.2   & 0.0125 & \textbf{1}    &       8.7    \\
 ML-II GP             & 0.0843 & 0.82 &       3.8 & 0.1321 &  0.77 & 3.9  & 0.1817 &0.72 &       4     \\
 GAM SS LASSO         & 0.2749 & 1    &     386.7 & 0.3563 &  0.93 & 390.1  & 0.4831 & 0.82 &     393.6    \\
 GAM LASSO            & 0.0725 & 0.53 &       3.8& 0.1373 &  0.43 & 3.9 & 0.1849 & 0.4  &       4.1      \\
 LASSO                & 0.3777 & 0.44 &       \textbf{0.3}& 0.4796 &  0.35 & \textbf{0.3} & 0.6254 & 0.3  &       \textbf{0.3}      \\
 MCP                  & 0.2731 & 0.64 &       0.3& 0.3755 &  0.54 & 0.3  & 0.4815 & 0.49 &       0.4     \\
\hline
\end{tabular}
    \caption{Experiment 1 performance percentiles for implemented models. Bold indicates best performing method per column. Note that the percentiles are reported such that the $75^{th}$ percentile performance is worse than the $25^{th}$ percentile performance, regardless of the metric. All SSVGP implementations outperformed the comparator set in variable selection and predictive accuracy for all percentiles reported.}
    \label{t_a1}
\end{table}

\begin{table}[H]
    \centering
\begin{tabular}{llll}
\hline
              & MSE             & MCC         & Runtime     \\
\hline
 SSVGP(m=n)   & 0.0616 ± 0.1267 & 0.85 ± 0.25 & 20.3 ± 1.2  \\
 SSVGP(m=n/2) & \textbf{0.0187} ± 0.0422 & \textbf{0.94} ± 0.16 & 11.0 ± 0.9  \\
 SSVGP(m=n/4) & 0.0258 ± 0.0543 & 0.93 ± 0.17 & 8.4 ± 0.6   \\
 ML-II GP     & 0.143 ± 0.0933  & 0.79 ± 0.13 & 3.9 ± 0.2   \\
 GAM SS LASSO   & 0.4174 ± 0.2154 & 0.87 ± 0.14 & 389.9 ± 4.8 \\
 GAM LASSO    & 0.164 ± 0.138   & 0.47 ± 0.1  & 4.0 ± 0.2   \\
 LASSO        & 0.5065 ± 0.1548 & 0.37 ± 0.1  & \textbf{0.3} ± 0.0   \\
 MCP          & 0.4103 ± 0.1885 & 0.58 ± 0.11 & 0.3 ± 0.0   \\
\hline
\end{tabular}
    \caption{Experiment 1 mean $\pm$ standard deviation performance for implemented models. MSE is mean squared error and MCC is Matthews Correlation Coefficient. Bold indicates best performing method per column. All SSVGP implementations outperformed the comparator set on average variable selection and predictive accuracy, with the exception that the GAM spike and slab lasso of \citet{bai2020spike} marginally outperfomed the SSVGP with $m=n$ on average with respect to variable selection accuracy (MCC).}    
    \label{t_a2}
\end{table}

\begin{table}[H]
    \centering
\begin{tabular}{llll}
\hline
              & MSE             & MCC         & Runtime (s)     \\
\hline
 Savitsky et al (2011) MCMC spike and slab GP (single trial)    & 0.0067 & 1 & $>$10000  \\
 
\hline
\end{tabular}
    \caption{Single trial performance of Savitsky et al's (2011) single trial of their MCMC spike and slab GP on the high-dimensional experiment. MSE is mean squared error and MCC is Matthews Correlation Coefficient.}
    \label{tab:my_label}
\end{table}

\subsection{Experiment 2}    

Here we test the scalability and predictive accuracy of our SSVGP in a large-scale sparse input prediction challenge. The aim of this experiment is to demonstrate that regardless of whether variable selection is a central task, our method can substantially improve on the predictive performance of benchmark scalable GP approximations by taking advantage of input relevance sparsity when available. In this experiment we draw a univariate response from the following interactive function of two inputs:
\begin{align}
y(\bm x) & = f(\bm x) + \epsilon \\
f(\bm x) & =  \tan(x_1)+\tan(x_2)+\sin(2\pi x_1)+\sin(2\pi x_2)+\cos(4\pi^2x_1x_2)+\tan(x_1x_2) \\
\epsilon & \sim \mc N(0,\sigma^2)  \\
(x_1,x_2) & \sim \mc U[0,1]^2
\end{align}
We then augment the dataset with $d-2$ noise dimensions $\bm z$, where $corr(x_i,z_j) = corr(z_j,z_k) = 0.5 \quad \forall i \in [2], \quad \forall  j,k \in [d-2]$. To generate this correlation structure we pass $\bm x$ through an inverse standard Gaussian CDF, draw $\bm z|\bm x$ using standard Gaussian conditioning formulae, and then pass $\bm x, \bm z$ back through a standard Gaussian CDF. We fix $nd =10^8$ and consider two variations in dimensions: (1) an ultra-high dimensional design $(n=10^4,d=10^4)$ and (2) a one million training point design $(n=10^6, d=10^2$). For testing we draw an equivalent dataset of size $n_* = 10^4$ using the same generative process, but now with $x_1,x_2$ as grid spaced over $[0,1]$. This enables a 2d prediction surface to be produced as a function of the two generating inputs and thus facilitates clear visual comparisons in results. 

We implement the SSVGP using the same implementation settings as previously but use a smaller base learning rate of $\eta = 0.01$. We use the smaller learning rate in large-$n$ designs when using minibatch sizes such that $m\ll n$, as we find this leads to more stable learning due to the high degree of gradient variance. The specific minibatch sizes we choose are $m \in \{64,128,256\}$. We constrain the nearest neighbours in the minibatch to be chosen from a random subset of $min(n,10^4)$ datapoints to limit the per-iteration nearest neighbour search cost. We nearest neighbour truncate the LOOPD using $\tilde m = 64$ neighbours, and the predictive distribution using $m_* = 256$ neighbours. We find $\tilde m \leq 64$ is required for competitive runtimes when $n=10^6$ and $m_*\geq256$ is required for (near) optimal predictive performance vs. using $m=n$. We consider $\tilde m = 64$ and $m_* = 256$ as default recommendations for our method. We use KD-trees \cite{bentley1975multidimensional} whenever the active input dimensionality is $d<100$, else we use Ball trees \cite{omohundro1989five}.

Since the experiment is focussed on predictive accuracy, we also do the following to speed up computation time: (i) We select the best model rather than averaging as this means only one set of GP predictive moments are computed, and (ii) We compress the model set such that if any $M$ models select the same variables, we keep only the model $i \in [M]$ with the best LOOPD on a random subset of $n=1000$ training points. This leads to substantial time savings when $n=10^6$, since as $n \to \infty$ LOOPD dominates computation time due to its superlinear complexity. These are default recommendations when predictive accuracy is the main concern.

The two competitors we use are the sparse Gaussian process (SGP) of \citet{titsias2009variational}, and the stochastic variational Gaussian process (SVGP) of \citet{hensman2013gaussian}, which are popular benchmarks in the scalable GP literature. We follow the implementations in \citet{wang2019exact, chen2020stochastic} (512/1024 inducing points, 100 iterations/epochs and an ADAM learning rate of 0.1/0.01 for the SGP/SVGP respectively). However we deviate from this in the following cases : (1) In the ultra-high-dimensional design we train for 200 iterations/epochs as performance was extremely uncompetitive after 100 iterations/epochs. (2) In the one-million training point design we use 25 epochs for the SVGP as we found this did not change predictive accuracy noticeably but enabled $4 \times $ savings in runtime (runtimes were not competitive for the SVGP when $n=10^6$ using 100 epochs otherwise). We also use natural gradients for the SVGP following the GPytorch documentation recommendations. We use the same initialisation as our SSVGP for shared parameters, and standardise inputs and the response as previously.

Mean and standard deviation performance results from 3 trials of each design are in Fig. \ref{f5} in the main text. We also display in Fig. \ref{f4} of the main text the average prediction surfaces as a function of the two generating inputs for each method, against the true (noiseless) latent test surface $\bm f_*$. Full results tables and a breakdown of the runtime source for our SSVGP are below in Tables \ref{t_a3}-\ref{t_a4}. Our SSVGPs all performed best on average and kept competitive runtimes in both designs (faster than both comparators when $n=10^6$ and faster than the SVGP when $n=10^4$). We note the relatively faster runtimes achieved by our method in the $n=10^6$ case do not reflect better complexity with respect to $n$, but rather that more models in the BMA ensemble selected the same variables and thus the collapsed the model set was smaller for LOOPD computation.

In the ultra-high-dimensional design ($n=d=10^4$), runtime is dominated by a-CAVI training time, and the larger minibatch size therefore substantially increases total time. However in the one-million training point design ($n=10^6,d=10^2$), the runtime is dominated by LOOPD computation due to its superlinear complexity. In this case, we actually find $m=256$ runs fastest on average, as it tended to prune (select) fewer variables per BMA ensemble member (recall LOOPD compute time is linear in the number of non-pruned variables $p_k$ per model $\mc M_k$ in the BMA ensemble).

\begin{table}[H]
    \centering
\begin{tabular}{lll}
\hline
& \multicolumn{2}{c}{$(n=10^4,d=10^4)$} \\
              & MSE   & Runtime (mins)   \\
\hline
 SGP(512)     & 0.682 ± 0.015        & \textbf{14.1} ± 0.3       \\
 SVGP(1024)   & 0.644 ± 0.005        & 154.8 ± 3.3      \\
 SSVGP(m=256) & \textbf{0.26} ± 0.002         & 42.6 ± 5.9       \\
 SSVGP(m=128) & 0.261 ± 0.003        & 24.3 ± 4.3       \\
 SSVGP(m=64)  & 0.262 ± 0.004        & 17.9 ± 2.9       \\
\hline
\end{tabular} 
\begin{tabular}{ll}
\hline
 \multicolumn{2}{c}{$(n=10^6,d=10^2)$} \\
 MSE   & Runtime (mins)   \\
\hline
 0.263 ± 0.002        & 288.7 ± 17.2     \\
 0.256 ± 0.002        & 365.6 ± 14.9     \\
 \textbf{0.251} ± 0.002        & \textbf{227.3} ± 23.2     \\
 0.251 ± 0.002        & 236.6 ± 15.6     \\
 0.251 ± 0.002        & 245.8 ± 17.0     \\
\hline
\end{tabular}
    \caption{Experiment 2 mean $\pm$ standard deviation results (MSE = mean squared error) from 3 trials on the two dataset sizes tested. Bold indicates the best performing method.}
    \label{t_a3}
\end{table}

\begin{table}[H]
    \centering
\begin{tabular}{llll}
\hline
 & \multicolumn{3}{c}{$(n=10^4,d=10^4)$} \\
              & Train   & LOOPD   & Test   \\
\hline
 SSVGP(m=256) & 35.4 ± 2.4      & 6.9 ± 3.5      & 0.2 ± 0.0   \\
 SSVGP(m=128) & 17.3 ± 0.7      & 6.7 ± 3.6      & 0.2 ± 0.0   \\
 SSVGP(m=64)  & 12.9 ± 0.7      & 4.8 ± 2.3      & 0.2 ± 0.0   \\
 \hline 
\end{tabular}
\begin{tabular}{lll}
\hline
 \multicolumn{3}{c}{$(n=10^6,d=10^2)$} \\
 Train   & LOOPD   & Test   \\
\hline
 11.9 ± 0.2      & 215.0 ± 23.2   & 0.4 ± 0.0   \\
 12.3 ± 0.2      & 223.9 ± 15.4   & 0.3 ± 0.0   \\
 12.1 ± 0.3      & 233.3 ± 16.7   & 0.3 ± 0.0   \\
\hline
\end{tabular}
    \caption{Experiment 2 mean $\pm$ standard deviation runtime (minutes) breakdown into training time (a-CAVI), LOOPD computation, and test time from 3 trials on the two dataset sizes tested.}
    \label{t_a4}
\end{table}

\subsection{Real Datasets from UCI Repository}
\begin{figure}[H]
    \centering
    \includegraphics[width  =1\textwidth]{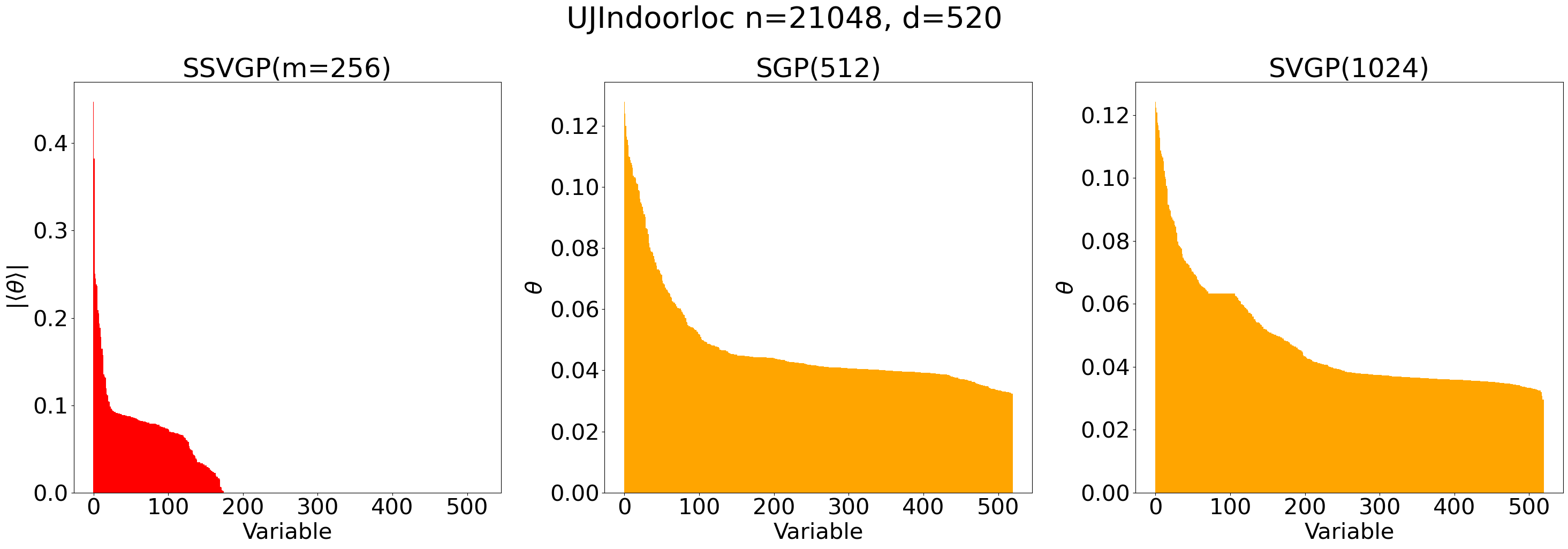} \\~\\
    \includegraphics[width  =1\textwidth]{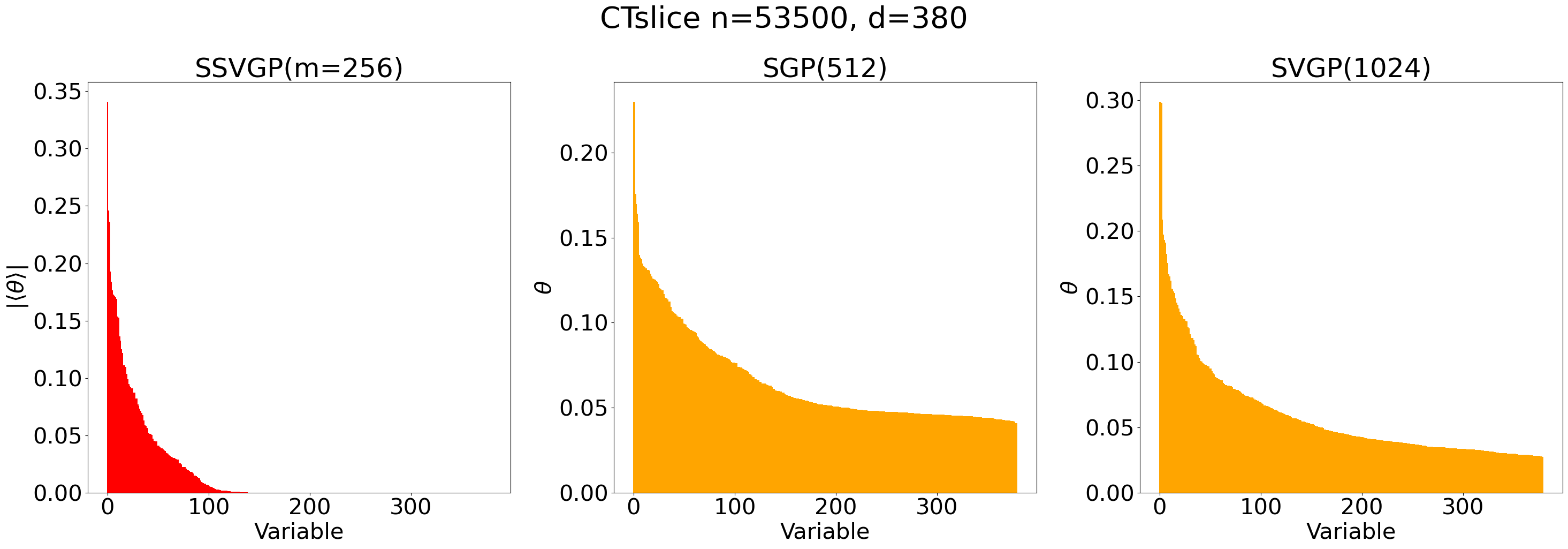}
    \caption{Size ordered inverse lengthscale profiles for our SSVGP with $m=256$ minibatching (posterior mean of $q_{\bm \psi}(\bm \theta)$ used), against sparse GP \cite{titsias2009variational} and stochastic variational GP \cite{hensman2013gaussian} on the real datasets. Since our SSVGP is parameterised such that $\bm \theta \in \mathbb R^d$, we plot $|\langle \bm \theta \rangle_{q_{\bm \psi}(\bm \theta)}|$. The median trial is used to construct the profiles, meaning that the $k^{th}$ column value is the median value of the $k^{th}$ largest inverse lengthscale recovered. We use the median to indicate how the profile looks for a `typical' trial, as the average is skewed by two trials with a larger number of selections. Our SSVGP learns an L0-sparse subset of inverse lengthscales through the BMA weighting over different models with varying aggressiveness in dropout pruning, but predicts with significantly more accuracy on average (see Table \ref{t_a5}).}
    \label{f_a3}
\end{figure}

We now test on two benchmark real datasets from the UCI repository with high-dimensional  inputs ($d\gg 100$): CTSLICE (n=53500, d=380) and UJINDOORLOC (n=21048, d=520). As variable selection accuracy cannot be verified here we focus again on the predictive accuracy and runtime of our method. We implement the SSVGP against the SGP and SVGP using exactly the implementation settings as in Experiment 2 but train the SGP and SVGP using the default 100 iterations/epochs and set the SSVGP's training minibatch to $m=256$ (the best performing in Experiment 2). We run these methods on 10 trials of random splits into training and test sets, with 4/5 used for training. As previously we standardise the data to have mean zero and unit variance. We also pass $\bm y$ through a Box-cox transformation due to the heavy skew in the distributions, to maximise consistency with the Gaussian modelling assumptions.  Fig. \ref{f6} in the main text contains the mean and standard deviation of MSE and runtimes obtained, and we display results tables below in \ref{t_a5}. We also highlight in Fig. \ref{f_a3} below the `sparsity' of the learned solutions by our SSVGP by comparison to the SGP/SVGP by comparing the recovered inverse lengthscales (post. mean used for SSVGP). 

\begin{table}[H]
    \centering
\begin{tabular}{llll}
\hline
& \multicolumn{3}{c}{CTSLICE (n=53500,d=385)} \\
              & MSE             & Runtime    & Variables used   \\
\hline
 SSVGP(m=256) & \textbf{0.003} ± 0.001 & 34 ± 2 & \textbf{142} ± 25    \\
 SGP(512)     & 0.006 ± 0.000 & \textbf{3} ± 0  & 385        \\
 SVGP(1024)   & 0.005 ± 0.000 & 60 ± 1 & 385        \\
\hline
\end{tabular}
\begin{tabular}{llll}
\hline
 \multicolumn{3}{c}{UJINDOORLOC (n=21048,d=520)} \\
               MSE             & Runtime    & Variables used   \\
\hline
 \textbf{0.027} ± 0.003 & 18 ± 0 & \textbf{170} ± 32     \\
0.038 ± 0.003 & \textbf{1} ± 0  & 520          \\
0.035 ± 0.003 & 25 ± 0 & 520        \\
\hline
\end{tabular}
\caption{Real dataset results for our for our SSVGP with $m=256$ minibatching, against SGP \cite{titsias2009variational} and SVGP \cite{hensman2013gaussian} on the real datasets. `Variables used' is measured by the number of dimensions $j=(1....J)$ which affect predictions when $x_{*(j)}$ varies. Bold indicates the best performing method per column.}
\label{t_a5}
\end{table}

\clearpage
%\bibliography{references}